\documentclass{article}

\usepackage{microtype}
\usepackage{graphicx}
\usepackage{subcaption}
\usepackage{booktabs} 

\usepackage{hyperref}

\usepackage[preprint]{icml2026}

\usepackage{amsmath}
\usepackage{amssymb}
\usepackage{mathtools}
\usepackage{amsthm}

\usepackage[capitalize,noabbrev]{cleveref}

\theoremstyle{plain}

\theoremstyle{definition}

\theoremstyle{remark}

\usepackage[textsize=tiny]{todonotes}

\icmltitlerunning{AERO: Autonomous Evolutionary Reasoning Optimization via Endogenous Dual-Loop Feedback}
\usepackage{epigraph}
\usepackage{multirow}
\usepackage[table,xcdraw]{xcolor}
\definecolor{RoyalBlue}{RGB}{65,105,225}
\usepackage[normalem]{ulem}
\useunder{\uline}{\ul}{}
\usepackage{tcolorbox}
\tcbuselibrary{skins, breakable, listings}
\newtcolorbox{promptBox}[1]{
    title=\textbf{#1},
    breakable,
    enhanced,
    fonttitle=\bfseries,
    colframe=RoyalBlue,
    boxrule=1pt,
    toprule=3pt,
    rounded corners,
    arc=2pt,
    top=1mm,bottom=1mm,left=1mm,right=1mm
}                                 
\usepackage{enumitem}
\begin{document}

\twocolumn[
  \icmltitle{AERO: \underline{A}utonomous \underline{E}volutionary \underline{R}easoning \underline{O}ptimization via Endogenous Dual-Loop Feedback}



  \icmlsetsymbol{equal}{*}

  \begin{icmlauthorlist}
    \icmlauthor{Zhitao Gao}{xjtu-cs,xjtu-moe}
    \icmlauthor{Jie Ma}{xjtu-cyber,xjtu-moe}
    \icmlauthor{Xuhong Li}{baidu}
    \icmlauthor{Pengyu Li}{xjtu-cs,xjtu-bdkelab}
    \icmlauthor{Ning Qu}{xjtu-cs,xjtu-bdkelab}
    \icmlauthor{Yaqiang Wu}{xjtu-cs,lenovo}
    \icmlauthor{Hui Liu}{lenovo}
    \icmlauthor{Jun Liu}{xjtu-cs,xjtu-moe,xjtu-bdkelab}
  \end{icmlauthorlist}

  \icmlaffiliation{xjtu-cs}{School of Computer Science and Technology, Xi’an Jiaotong University, Xi’an, 710049, China}
  \icmlaffiliation{xjtu-cyber}{School of Cyber Science and Engineering, Xi’an Jiaotong University, Xi’an, Shaanxi 710049, China}
  \icmlaffiliation{xjtu-moe}{MOE KLINNS Lab, Xi’an Jiaotong University, Xi’an, 710049, China}
  \icmlaffiliation{xjtu-bdkelab}{Shaanxi Province Key Laboratory of Big Data Knowledge Engineering, Xi’an Jiaotong University, Xi’an, 710049, China}
  \icmlaffiliation{lenovo}{Lenovo AI Technology Center, CTOO, Lenovo}
  \icmlaffiliation{baidu}{Baidu Inc.}

  \icmlcorrespondingauthor{Jie Ma}{jiema@xjtu.edu.cn}

  \icmlkeywords{Machine Learning, ICML}

  \vskip 0.3in
]



\printAffiliationsAndNotice{}  

\begin{abstract}
Large Language Models (LLMs) have achieved significant success in complex reasoning but remain bottlenecked by reliance on expert-annotated data and external verifiers. While existing self-evolution paradigms aim to bypass these constraints, they often fail to identify the optimal learning zone and risk reinforcing collective hallucinations and incorrect priors through flawed internal feedback. To address these challenges, we propose \underline{A}utonomous \underline{E}volutionary \underline{R}easoning \underline{O}ptimization (AERO), an unsupervised framework that achieves autonomous reasoning evolution by internalizing self-questioning, answering, and criticism within a synergistic dual-loop system. Inspired by the \textit{Zone of Proximal Development (ZPD)} theory, AERO utilizes entropy-based positioning to target the ``solvability gap'' and employs Independent Counterfactual Correction for robust verification. Furthermore, we introduce a Staggered Training Strategy to synchronize capability growth across functional roles and prevent curriculum collapse. Extensive evaluations across nine benchmarks spanning three domains demonstrate that AERO achieves average performance improvements of 4.57\% on Qwen3-4B-Base and 5.10\% on Qwen3-8B-Base, outperforming competitive baselines. Code is available at \url{https://github.com/mira-ai-lab/AERO}.

\end{abstract}
\section{Introduction}
\begin{figure}[tbp]
    \centering
    \includegraphics[width=0.9\linewidth]{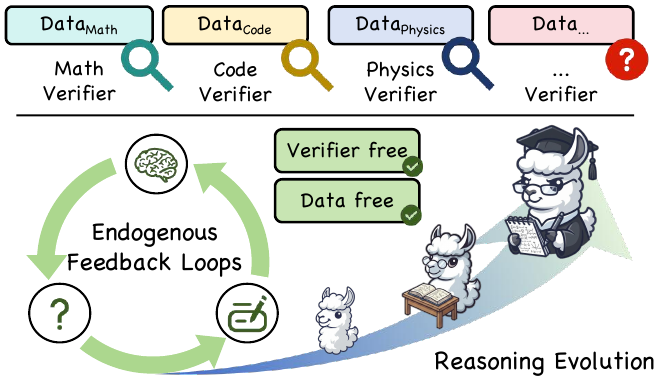}
    \caption{The AERO framework for unsupervised self-evolution. AERO internalizes three core capabilities, which include self-questioning, self-answering, and self-criticism, within a unified dual-loop system to enable autonomous growth without any reliance on external data or verifiers.}
    \label{fig:challenge}
    \vskip -0.1in
\end{figure}
Recently, Large Language Models (LLMs) have demonstrated strong reasoning capabilities~\cite{jaech2024openai, guo2025deepseek, ma2025deliberation}. While paradigms like Reinforcement Learning with Verifiable Rewards (RLVR) have driven much of this progress~\cite{lambert2024tulu}, they remain dependent on expert-level queries and automated verifiers such as code compilers or math engines~\cite{zhao2025absolute, huang2025formarl}. This dependency creates a bottleneck that restricts model growth to the limits of predefined data and prevents the discovery of reasoning patterns that exceed existing human knowledge~\cite{yuan2024self}. To overcome these constraints, the \textbf{Self-Evolution} paradigm has emerged as a critical pathway toward achieving higher intelligence~\cite{tan2024large, tao2024survey}. By allowing models to improve iteratively through learning from their own generated data and experiences, this shift transforms LLMs from passive recipients of information into active participants in their own developmental cycle~\cite{huang2025r, kuba2025language}.

Despite their potential to decouple model training from external data and supervision, existing Self-Evolving paradigms face two fundamental limitations:
1) Current mechanisms often cause the model to fall into a \textbf{sub-optimal learning zone} because they lack a clear way to adjust task difficulty. This sub-optimal zone occurs when task generation misses the \textit{solvability gap}, which represents the area where tasks are neither too simple to provide new insights nor too hard for the model to understand~\cite{zhang2025interplay}. Without a strategy to target this gap, the model faces a significant loss in learning efficiency. It either stops progressing by repeating what it already knows or fails to learn because it faces overly complex problems that produce feedback no better than random noise~\cite{kuba2025language, zhao2025absolute}.
2) To replace external verifiers, existing paradigms typically rely on internal indicators such as majority voting~\cite{he2025visplay, huang2025r} and decoding confidence~\cite{liu2025nover, yu2025rlpr, zhou2025reinforcing} to provide reward signals. These methods work on the assumption that agreement or high probability is the same as logical correctness. However, when a model holds an incorrect belief, these indicators become unreliable and instead \textbf{reinforce collective hallucinations and incorrect priors}. This traps the model in a feedback loop that confirms its own mistakes, which eventually drives the learning process away from logical truth.

To address these challenges, we propose \textbf{Autonomous Evolutionary Reasoning Optimization (AERO)}, an unsupervised framework structured as an inner-outer dual-loop system that internalizes three synergistic capabilities within a single LLM: Self-Questioning (Generator), Self-Answering (Solver), and Self-Criticism (Refiner), as illustrated in Figure~\ref{fig:challenge}.
To prevent the LLM from falling into sub-optimal learning zones, AERO is inspired by the theory of the \textbf{\textit{Zone of Proximal Development (ZPD)}}~\cite{vygotsky1978mind}, which posits that cognitive development is maximized when task difficulty is precisely calibrated to the learner's current reasoning capabilities. We operationalize this principle by defining task difficulty through the LLM's level of reasoning uncertainty. Specifically, tasks that exhibit a moderate degree of uncertainty for the current LLM signify the ideal ``solvability gap''. AERO utilizes normalized Shannon entropy to quantify this uncertainty, guiding the LLM to autonomously generate tasks that fall within its optimal learning zone at the frontier of its reasoning capabilities. 

To overcome the risk of reinforcing collective hallucinations and incorrect priors, we introduce \textbf{Independent Counterfactual Correction (ICC)}, which compels the LLM to reconstruct its reasoning path under the assumption that the initial reasoning was flawed. By requiring the answer convergence of independent paths rather than mere statistical consensus, ICC provides a high-reliability truth proxy for the policy optimization. In addition, to maintain evolutionary stability and prevent curriculum collapse, we implement a \textbf{Staggered Training Strategy} that synchronizes the capability growth of all functional roles. Evaluations across nine benchmarks spanning general reasoning, mathematical reasoning, and physical reasoning domains demonstrate that AERO consistently outperforms competitive baselines. Specifically, it achieves average performance gains of 4.57\% on Qwen3-4B-Base and 5.10\% on Qwen3-8B-Base, while maintaining a consistent improvement trend across diverse architectures.

The primary contributions of this work are as follows:
\begin{itemize}
    \item We propose the Autonomous Evolutionary Reasoning Optimization (AERO) dual-loop framework, which achieves autonomous self-evolution without external verifiers or labels. This is the first framework to achieve the simultaneous evolution of Self-Questioning, Self-Answering, and Self-Criticism within a single LLM, enabling a comprehensive reasoning evolution.
    \item We introduce entropy-based Zone of Proximal Development positioning to target the optimal learning zone and Independent Counterfactual Correction to provide highly reliable logical verification. Additionally, we propose a Staggered Training Strategy to mitigate curriculum collapse and stabilize the resulting evolutionary dynamics.
    \item We conduct extensive evaluations across nine benchmarks to verify the effectiveness and superiority of AERO. Futhermore, we demonstrate its robustness and evolutionary stability in sustaining continuous growth, proving that reasoning capabilities can grow effectively through purely endogenous feedback.
\end{itemize}

\section{Methodology}
\begin{figure*}[tbp]
    \centering
    \includegraphics[width=0.9\linewidth]{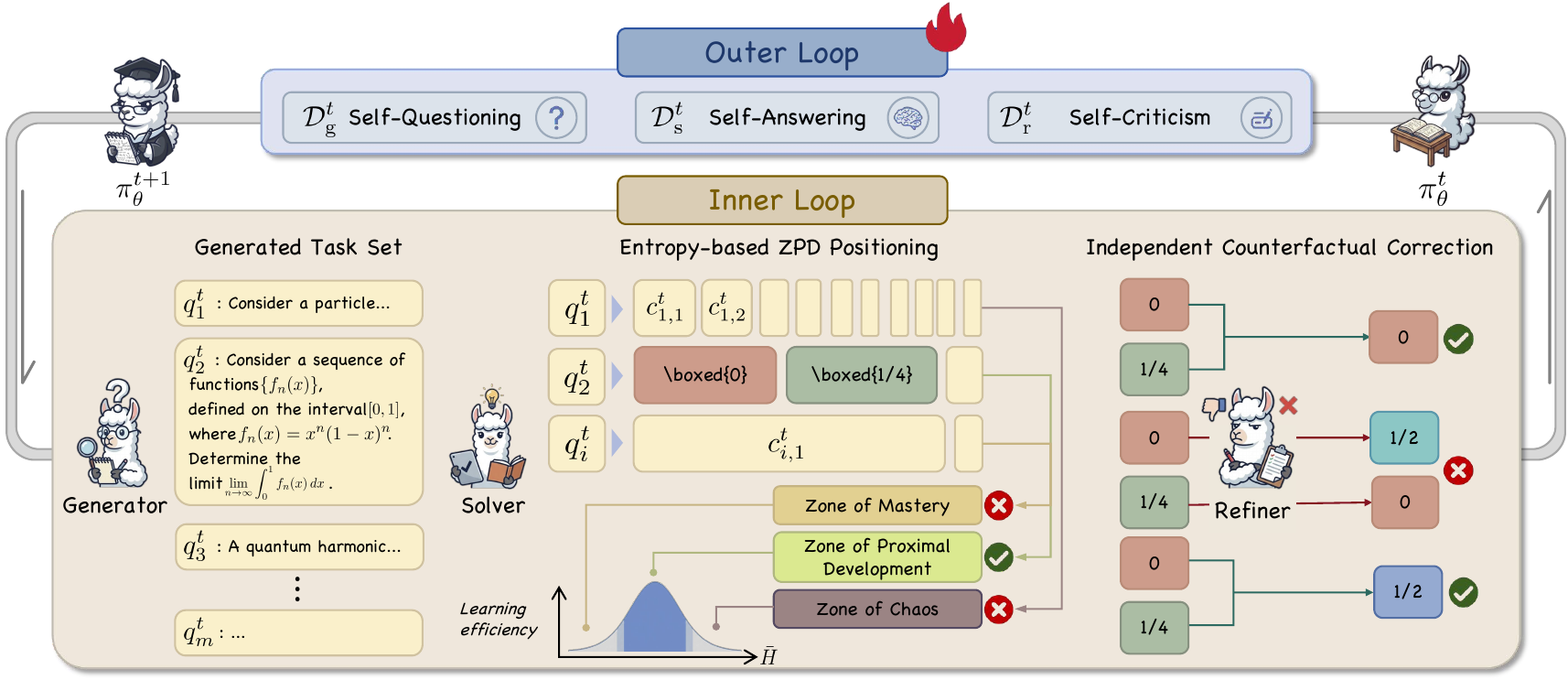}
    \caption{The AERO framework consists of an inner loop for autonomous experience synthesis and an outer loop for preference-based policy optimization. Within the inner loop, the single model adopts generator, solver, and refiner roles to produce tasks and reasoning trajectories. These verified experiences are then utilized in the outer loop for policy update.}
    \label{fig:arc}
\end{figure*}
\subsection{Framework Overview}
The AERO framework empowers a single LLM, denoted as $\pi_\theta$, to autonomously evolve its reasoning capabilities through a dual-loop architecture without external supervision or human-annotated data. As illustrated in Figure~\ref{fig:arc}, the system is composed of an inner loop for experience synthesis and an outer loop for preference-based policy optimization. 

The evolution of $\pi_\theta$ proceeds in an iterative manner. At round $t$, the inner loop sees the model $\pi_{\theta}^t$ operate as an autonomous data factory to synthesize a preference dataset $\mathcal{D}^t$. In the outer loop, $\mathcal{D}^t$ is leveraged to update the model parameters from $\pi_\theta^t$ to $\pi_\theta^{t+1}$ via preference optimization, thereby internalizing the specialized capabilities of the three roles. AERO realizes an automated curriculum that progressively shifts the optimal learning zone toward higher complexity, driving a steady advancement of the LLM's reasoning capabilities through continuous self-evolution.

\subsection{Inner Loop}
The inner loop functions as a self-play sandbox where the Generator ($\pi_\text{g}^t$), Solver ($\pi_\text{s}^t$), and Refiner ($\pi_\text{r}^t$) collaborate to synthesize verified experiences in round $t$. Crucially, these three roles represent distinct functional capacities of $\pi_\theta^t$, which are activated through specific task-oriented prompts in Appendix~\ref{sec: prompt}. This process is driven by two synergistic mechanisms: Entropy-based ZPD positioning and ICC-based logical verification.

\subsubsection{Entropy-based ZPD Positioning}
To identify the reasoning capability frontier where learning is most effective, we implement a selection mechanism inspired by the theory of ZPD~\cite{vygotsky1978mind}. In round $t$, the Generator $\pi_\text{g}^t$ first synthesizes a set of $m$ challenging, competition-level reasoning tasks $\mathcal{Q}^t = \{q_1^t, \dots, q_m^t\}$ across various academic domains. For each specific task $q_i^t \in \mathcal{Q}^t$, the Solver $\pi_\text{s}^t$ generates $n$ independent reasoning trajectories $\mathcal{Y}_{i}^t = \{y_{i,1}^t, \dots, y_{i,n}^t\}$.
The final answers are extracted from these $n$ trajectories and grouped into $k$ unique clusters $\mathcal{C}_i^t = \{c_{i,1}^t, \dots, c_{i,k}^t\}$ based on their semantic equivalence, the specific implementation of which is detailed in Appendix~\ref{clustering_details}.

We employ \textbf{Shannon entropy} as the diagnostic metric to measure the uncertainty within the probability distribution formed by the reasoning trajectories of model $\pi_{\theta}^t$. By treating the normalized frequencies of answer clusters as a distribution, Shannon entropy allows us to quantify the degree of reasoning uncertainty, which reflects the difficulty level of task $q_i^t$ relative to the current model $\pi_{\theta}^t$. Specifically, we define the Normalized Shannon Entropy $\bar{H}(q_i^t)$ as:
\begin{equation}
    \bar{H}(q_i^t) = - \frac{1}{\log_2 n} \sum_{j=1}^{k} P(c_{i,j}^t) \log_2 P(c_{i,j}^t),
\end{equation}
where $P(c_{i,j}^t) = |c_{i,j}^t| / n$ denotes the empirical frequency of the $j$-th answer cluster for task $q_i^t$ in round $t$. The normalization factor $1/\log_2 n$ ensures that $\bar{H}(q_i^t)$ remains within the interval $[0, 1]$, where a value of 0 indicates total consensus and 1 represents maximum divergence.

By mapping these entropy values to the cognitive landscape, we categorize each task $q_i^t$ into three distinct regions based on the current reasoning capability of the model: 

\textbf{Zone of Mastery} ($\bar{H}(q_i^t) < \tau_{low}$): Tasks falling within this range are those where high consensus indicates the required logic is already internalized by $\pi_\theta^t$, which offers a negligible learning gradient for further policy evolution.

\textbf{Zone of Proximal Development} ($\tau_{low} \le \bar{H}(q_i^t) \le \tau_{high}$): This is the optimal learning zone where moderate reasoning uncertainty identifies the solvability gap. These tasks are most conducive to the cognitive growth of the model.

\textbf{Zone of Chaos} ($\bar{H}(q_i^t) > \tau_{high}$): This zone represents tasks that are far too difficult for the model $\pi_{\theta}^t$'s reasoning capabilities. When faced with overwhelming complexity, the LLM produces highly random and inconsistent guesses which act as ``noise'' that can confuse itself and lead to training instability.

AERO focuses exclusively on the \textit{Zone of Proximal Development} for the subsequent stages of the framework. This filtering process ensures that the verification stage targets only those data points providing the most productive learning signals for policy optimization, thereby maintaining an efficient and focused evolutionary trajectory.

\subsubsection{ICC-based Logical Verification}
For each task $q_i^t$ positioned within the ZPD, we establish truth proxies through ICC. Traditional verification methods, such as majority voting or decoding confidence, often fail in self-evolution scenarios because they risk reinforcing collective hallucinations and incorrect priors. ICC addresses this limitation by utilizing logical convergence under counterfactual pressure to verify reasoning correctness without requiring external gold labels.

The process begins by identifying the two most frequent answer clusters, $c_{i,1}^t$ and $c_{i,2}^t$, generated by the Solver $\pi_\text{s}^t$ for task $q_i^t$, which represent the model $\pi_{\theta}^t$'s primary competing consensuses. The Refiner $\pi_\text{r}^t$ is then prompted to re-solve the task $q_i^t$ while operating under the counterfactual assumption that the previously proposed solution is incorrect. This constraint breaks the cycle of confirmation bias, forcing the LLM to rethink the task and construct an independent reasoning path to verify the correct solution.

The correction path starting from cluster $c_{i,j}^t$, where $j \in \{1, 2\}$, is reoresented as $\tilde{y}_{i,j}^t$. We define the formal convergence condition as:
\begin{equation}
    \text{res}(\tilde{y}_{i,1}^t) = \text{res}(\tilde{y}_{i,2}^t),
\end{equation}
where the function $\text{res}(\cdot)$ extracts the final answer from a reasoning trajectory. The reasoning path $\tilde{y}_{i,1}^t$ is established as a verified truth proxy $\tilde{y}_{i}^t$ if this equality holds. Otherwise, if the correction trajectories fail to yield consistent results, the task is considered unresolved and is discarded from the synthesis of training datasets for both the Solver and Refiner roles.

\subsubsection{Tri-role Preference Synthesis}
The inner loop ends with transforming verified experiences from round $t$ into three specialized preference datasets: $\mathcal{D}_{\text{g}}^t$, $\mathcal{D}_{\text{s}}^t$, and $\mathcal{D}_{\text{r}}^t$. 
To facilitate preference optimization, we map synthesized experiences to binary labels $z \in \{0, 1\}$, where $z=1$ indicates a \textit{chosen} output and $z=0$ denotes a \textit{rejected} one.

For the Generator, $\mathcal{D}_{\text{g}}^t$ utilizes all $m$ generated tasks to help the LLM identify its reasoning frontier. We use the indicator function $\mathbb{I}_{\text{ZPD}}(q_i^t)$, which equals 1 if a task falls within the Zone of Proximal Development and 0 otherwise:
\begin{equation}
    \mathcal{D}_{\text{g}}^t = \left\{ \big(q_i^t, \mathbb{I}_{\text{ZPD}}(q_i^t)\big) \right\}_{i=1}^m.
    \label{eq:d_g}
\end{equation}

For the Solver, $\mathcal{D}_{\text{s}}^t$ is synthesized by evaluating initial reasoning trajectories against the ICC-verified truth proxy $\tilde{y}_{i}^t$ based on their result equivalence:
\begin{equation}
\begin{aligned}
    \mathcal{D}_{\text{s}}^t = \big\{ & (q_i^t, y_{i,j}^t, \mathbb{I}[\text{res}(y_{i,j}^t) = \text{res}(\tilde{y}_{i}^t)]) \\
    & \mid \mathbb{I}_{\text{ZPD}}(q_i^t)=1, 1 \le i \le m, 1 \le j \le n \big\}.
    \label{eq:d_s}
\end{aligned}
\end{equation}

For the Refiner, $\mathcal{D}_{\text{r}}^t$ captures the self-correction process by extracting trajectories that transition from a flawed state to a verified one. Specifically, we retain correction paths $\tilde{y}_{i,j}^{t}$ as positive samples only if the initial cluster result is incorrect, yet the subsequent refinement successfully reaches the truth proxy:
\begin{equation}
\begin{aligned}
    \mathcal{D}_{\text{r}}^t = \big\{ & (q_i^t, c_{i,j}^t, \tilde{y}_{i,j}^t, 1) \mid \mathbb{I}_{\text{ZPD}}(q_i^t)=1, \\
    & \text{res}(c_{i,j}^t) \neq \text{res}(\tilde{y}_{i}^t) \land \text{res}(\tilde{y}_{i,j}^t) = \text{res}(\tilde{y}_{i}^t), \\
    & 1 \le i \le m, j \in \{1, 2\} \big\}.
    \label{eq:d_r}
\end{aligned}
\end{equation}
Crucially, the datasets for the Solver and Refiner are constructed exclusively from the subset of tasks that satisfy the ZPD criteria, ensuring the LLM learns from the most productive signals.

By constructing these datasets independently, the framework prepares the necessary signals for the subsequent outer loop optimization. This decoupled organization is essential for the Staggered Training Strategy, a mechanism we discuss in detail in Section~\ref{sec:staggered}. 

\subsection{Outer Loop} \label{sec:outer_loop}
The outer loop translates the synthesized experiences into policy updates by optimizing $\pi_{\theta}$ across its three functional roles. To ensure a stable evolutionary trajectory, we introduce a temporal decoupling mechanism that synchronizes the growth of different capabilities.
\subsubsection{Staggered Training Strategy} \label{sec:staggered}
\begin{figure}[t]
    \centering
    \includegraphics[width=0.6\linewidth]{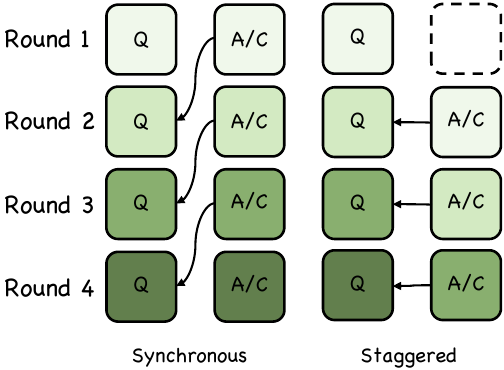}
    \caption{Synchronous vs. Staggered Training Strategy. Colors indicate the progression of training rounds. Arrows illustrate the capability alignment between Self-Questioning (Q) and Self-Answering/Self-Criticism (A/C). While synchronous training leads to capability asynchrony, the staggered approach introduces a temporal offset to synchronize growth across all functional capabilities.}
    \label{fig:Staggered}
    \vskip -0.1in
\end{figure}

A major limitation in LLM self-evolution is \textit{curriculum collapse}, where performance stops improving or even declines during iterative self-play~\cite{huang2025r, jiang2025bootstrapping, wang2025space}. We attribute one of the causes of this instability to capability asynchrony, where the learning speed of Self-Answering and Self-Criticism capabilities exceeds that of Self-Questioning. Specifically, as shown in Figure~\ref{fig:Staggered}, under the standard synchronous training strategy, $\pi_{\theta}^{t}$ masters its round $t$ ZPD tasks after the outer-loop parameter update. However, during round $t+1$, the newly synthesized ZPD tasks remain anchored to the capability of $\pi_{\theta}^{t-1}$ because the diagnostic ZPD signals are derived from the responses of $\pi_{\theta}^{t-1}$. Consequently, these tasks have already entered the Zone of Mastery for the updated model $\pi_{\theta}^{t}$, leading to vanishing learning gradients and subsequent training failure.

To mitigate this asynchrony, we introduce the \textbf{Staggered Training Strategy}, a simple but effective approach designed to synchronize the advancement of role-specific capabilities. This mechanism creates a temporal offset by using current Self-Questioning data alongside historical Self-Answering and Self-Criticism data to ensure the curriculum remains challenging. The training dataset $\mathcal{D}_{\text{total}}^t$ at round $t$ is formally organized as follows:
\begin{equation}
    \mathcal{D}_{\text{total}}^t = 
    \begin{cases} 
        \mathcal{D}_{\text{g}}^t, & t = 1, \\
        \mathcal{D}_{\text{g}}^t \cup \mathcal{D}_{\text{s}}^{t-1} \cup \mathcal{D}_{\text{r}}^{t-1}, & t > 1.
    \end{cases}
\end{equation}
By implementing this staggered data flow, the AERO framework effectively prevents curriculum collapse. This strategy ensures that the tasks generated in each round consistently target the updated Zone of Proximal Development of the model, which maintains a steady evolutionary pressure and drives a continuous capability advancement. We provide detailed empirical evidence for the efficacy of this strategy in Section~\ref{sec:analysis}.
\subsubsection{KTO-based Optimization and Evolutionary Dynamics}
The Staggered Training Strategy requires an optimization algorithm capable of handling binary preference signals while supporting stable offline updates across decoupled datasets. We adopt Kahneman-Tversky Optimization (KTO)~\cite{ethayarajh2024kto} to fulfill these requirements. KTO maximizes the expected utility of generation outputs based on the human decision-making model proposed by Kahneman and Tversky. In each round $t$, we optimize the current policy $\pi_\theta$ using the model from the previous iteration $\pi_\theta^t$ as a fixed reference policy $\pi_{\text{ref}}$. The objective is to minimize the KTO loss $\mathcal{L}_{\text{KTO}}$ to obtain the updated parameters for $\pi_\theta^{t+1}$:
\begin{equation}
    \mathcal{L}_{\text{KTO}}(\pi_\theta, \pi_\theta^t) = \mathbb{E}_{(x,y) \sim \mathcal{D}_{\text{total}}^t} \left[ \lambda_y - v(x, y) \right],
    \label{eq:kto_loss}
\end{equation}
where the definitions of input $x$ and output $y$ are specific to the functional role being optimized. The value function $v(x, y)$ models human utility perception as follows:
\begin{equation}
v(x, y) = 
    \begin{cases} 
        \lambda_{\text{p}} \sigma\left( \beta(r_\theta(x, y) - z_0) \right) & \text{if } y \sim Y_{\text{p}} \mid x, \\ 
        \lambda_{\text{n}} \sigma\left( \beta(z_0 - r_\theta(x, y)) \right) & \text{if } y \sim Y_{\text{n}} \mid x. 
    \end{cases}
\end{equation}
In this formulation, $r_\theta(x, y) = \log (\pi_\theta(y|x) / \pi_\theta^t(y|x))$ represents the implied reward relative to the reference model $\pi_\theta^t$ from the previous iteration. The reference point $z_0$ is defined as the KL divergence between $\pi_\theta$ and $\pi_\theta^t$, while $\beta$ modulates the risk aversion of the model. 

KTO is particularly suitable for the AERO framework for two primary reasons. First, it operates directly on binary labels rather than paired comparisons, which allows for efficient optimization despite the skewed distributions of chosen versus rejected trajectories within $\mathcal{D}_{\text{total}}^t$. Second, the offline nature of KTO is inherently suited for our Staggered Training Strategy because it enables stable policy updates using historical data from round $t-1$ without the need for active on-policy sampling.

\begin{figure}[tbp]
    \centering
    \includegraphics[width=0.7\linewidth]{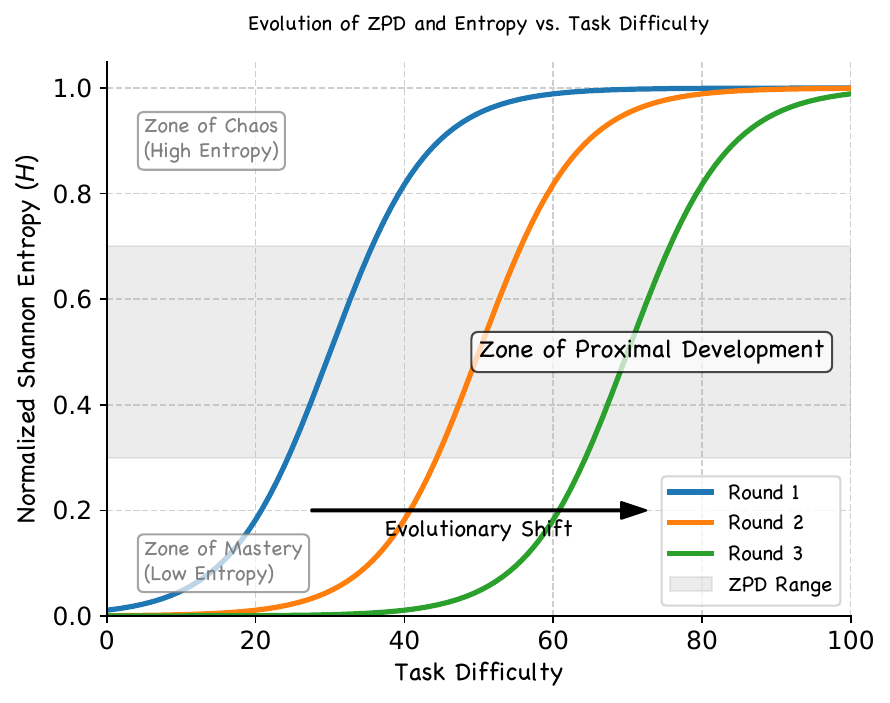}
    \caption{Evolution of ZPD and response entropy across iterations. The rightward shift of the curves from Round 1 to Round 3 demonstrates the model's cognitive growth, as the ZPD dynamically advances toward higher difficulty levels.}
    \label{fig:ZPD}
    \vskip -0.1in
\end{figure}

Algorithm~\ref{alg:aero} details the dual-loop optimization procedure. This optimization objective drives the continuous self-evolutionary process, which we conceptually illustrate in  Figure~\ref{fig:ZPD}. As the uncertainty curves ($\bar{H}(q)$) shift rightward across rounds, tasks once in the \textit{Zone of Chaos} move into the \textit{Zone of Proximal Development}, while previous ZPD tasks transition into the \textit{Zone of Mastery}. This shift demonstrates how AERO effectively advances the LLM's reasoning frontier in every iteration.
\begin{table*}[tbp]
\centering
\small
\caption{Main Results Comparison. Performance comparison of AERO against competitive self-evolution baselines, including R-Zero~\cite{huang2025r} and Absolute Zero~\cite{zhao2025absolute} across nine benchmarks in three reasoning domains. For each base model, \textbf{bold} denotes the best performance and \underline{underline} represents the second-best result. The grey-shaded rows (R5) represent the final evolutionary state of the AERO framework after five rounds of dual-loop optimization. The symbol ``-'' indicates that the results are not reported in the original papers of the respective baselines.}
\label{tab:main_v2}
\setlength{\tabcolsep}{3pt}
\begin{tabular}{@{}lccccccccc@{}}
\toprule
\multicolumn{1}{c}{}                                      & \multicolumn{3}{c}{\textbf{Mathematical Reasoning}} & \multicolumn{3}{c}{\textbf{Physical Reasoning}}               & \multicolumn{3}{c}{\textbf{General Reasoning}}           \\ \cmidrule(l){2-10} 
\multicolumn{1}{c}{\multirow{-2}{*}{\textbf{Model Name}}} & \textbf{GSM8K}  & \textbf{MATH500}  & \textbf{AMC}  & \textbf{UGPhysics} & \textbf{PhysicsEval} & \textbf{PHYBench} & \textbf{SuperGPQA} & \textbf{MMLU-Pro} & \textbf{GPQA-D} \\ \midrule
Qwen3-4B-Base                                             & 87.8            & 68.2              & 47.5          & 12.8               & 79.8                 & 2.7               & 25.4               & 51.6              & 26.3            \\
+ R-Zero                                                  & {\ul 92.1}      & {\ul 74.8}        & 48.2          & -                  & -                    & -                 & \textbf{27.8}      & 54.2              & \textbf{36.4}   \\
+ AZR                                                     & 89.3            & \textbf{76.2}     & 50.0          & -                  & -                    & -                 & 27.1               & {\ul 56.2}        & {\ul 35.3}      \\
+ AERO R1                                                  & 87.9            & 70.8              & 49.3          & 13.2               & 79.5                 & 2.7               & 26.3               & 52.4              & 26.3            \\
+ AERO R2                                                  & 90.2            & 72.6              & 46.7          & 15.3               & 80.1                 & 2.7               & 26.7               & 53.3              & 32.3            \\
+ AERO R3                                                  & 91.8            & 73.8              & 51.5          & 16.2               & {\ul 80.3}           & 3.4               & 27.1               & 53.9              & 33.3            \\
+ AERO R4                                                  & {\ul 92.1}      & 74.4              & {\ul 53.0}    & {\ul 18.5}         & \textbf{80.6}        & {\ul 3.7}         & 27.4               & 55.6              & 34.3            \\
\rowcolor[HTML]{EFEFEF} 
+ AERO R5                                                  & \textbf{92.4}   & {\ul 74.8}        & \textbf{54.5} & \textbf{19.4}      & 79.3                 & \textbf{3.9}      & {\ul 27.6}         & \textbf{56.9}     & 34.3            \\ \midrule
Qwen3-8B-Base                                             & 89.1            & 78.0              & 52.0          & 13.2               & 86.2                 & 3.8               & 28.3               & 58.0              & 33.3            \\
+ R-Zero                                                  & 94.1            & {\ul 82.0}        & 61.7          & -                  & -                    & -                 & 31.4               & 61.6              & \textbf{40.5}   \\
+ AZR                                                     & 92.0            & 76.6              & {\ul 62.5}    & -                  & -                    & -                 & \textbf{33.5}      & {\ul 62.5}        & 36.8            \\
+ AERO R1                                                  & 92.0            & 79.4              & 54.5          & 13.2               & 85.6                 & 3.4               & 30.0               & 59.3              & 33.8            \\
+ AERO R2                                                  & 92.9            & 79.2              & 56.7          & 14.8               & 86.1                 & 3.8               & 29.4               & 60.3              & 35.9            \\
+ AERO R3                                                  & 94.8            & 79.8              & 59.7          & 16.4               & 86.3                 & 4.0               & 31.1               & 60.3              & 34.3            \\
+ AERO R4                                                  & {\ul 95.8}      & 81.8              & 61.2          & {\ul 19.7}         & {\ul 86.9}           & {\ul 5.1}         & 32.1               & 61.5              & 38.4            \\
\rowcolor[HTML]{EFEFEF} 
+ AERO R5                                                  & \textbf{95.8}   & \textbf{82.2}     & \textbf{62.7} & \textbf{21.7}      & \textbf{87.9}        & \textbf{5.3}      & {\ul 32.5}         & \textbf{62.8}     & {\ul 36.9}      \\ \bottomrule
\end{tabular}
\end{table*}

\section{Experiments}
We conduct extensive experiments on nine benchmarks across three domains to address the following research questions: 
(1) Can AERO enable autonomous reasoning evolution on base models and outperform other baselines?
(2) Is AERO robust across different model families and parameter scales?
(3) How effective is each key component of AERO in driving its overall performance gains?
(4) Does the Staggered Training Strategy effectively prevent curriculum collapse?
(5) How reliable is the endogenous feedback generated by ICC in the absence of external ground-truth labels?
(6) Does AERO truly achieve automated curriculum learning throughout the multi-round self-evolution process?
\subsection{Experimental Setting}

\textbf{Implementation Details}.
Our AERO framework is implemented through an iterative process of experience synthesis and policy optimization. In each round of the inner loop, the Generator synthesizes $m=1,000$ tasks, while the Solver produces $n=16$ trajectories per task to compute the normalized Shannon entropy $\bar{H}(q_i^t)$. ZPD positioning is performed with thresholds $\tau_{low} = 0.3$ and $\tau_{high} = 0.7$. 
For the outer loop, policy optimization is conducted using KTO~\cite{ethayarajh2024kto}. The KL-divergence regularization coefficient $\beta$ is maintained at $0.1$. The detailed settings are provided in Appendix~\ref{exp_details}.

\textbf{Evaluation Benchmark}.
To comprehensively evaluate the reasoning capabilities of the AERO framework, we conduct experiments across nine challenging benchmarks spanning three distinct domains: 
(1) \textit{Mathematical Reasoning}, including GSM8K~\cite{cobbe2021training}, MATH500~\cite{hendrycks2measuring}, and AMC; 
(2) \textit{Physical Reasoning}, including UGPhysics~\cite{xuugphysics}, PhysicsEval~\cite{siddique2025physicseval}, and PHYBench~\cite{qiu2025phybench}; 
(3) \textit{General Reasoning}, including SuperGPQA~\cite{du2025supergpqa}, MMLU-Pro~\cite{wang2024mmlu}, and GPQA-Diamond~\cite{rein2024gpqa}. 
For all evaluations, we report the pass@1 accuracy under greedy decoding. Detailed specifications for each benchmark are provided in Appendix~\ref{benchmark}.

\textbf{Baseline Methods}.
We evaluate AERO against two competitive self-evolving baselines, R-Zero~\cite{huang2025r} and Absolute Zero~\cite{zhao2025absolute}. For a fair comparison, we utilize Qwen3-4B-Base and Qwen3-8B-Base~\cite{yang2025qwen3} as primary backbone models, ensuring strict consistency with the experimental settings used for all baseline methods. Additionally, we assess the generalization of AERO across a diverse set of instruction-tuned models, including Llama-3.2-3B-Instruct~\cite{dubey2024llama}, Qwen2.5-7B-Instruct, and Qwen2.5-32B-Instruct~\cite{qwen2025qwen25technicalreport}. 

To ensure a fair comparison, the performance metrics for R-Zero and Absolute Zero are cited directly from previous publications. As these baseline frameworks primarily focused on mathematical and general reasoning in their original evaluations, results for the physical reasoning domain are not available. We include these additional benchmarks further to demonstrate AERO's broad generalizability across diverse scientific domains.
\subsection{Main Results}
\begin{figure*}[h]
    \centering
    \includegraphics[width=\textwidth]{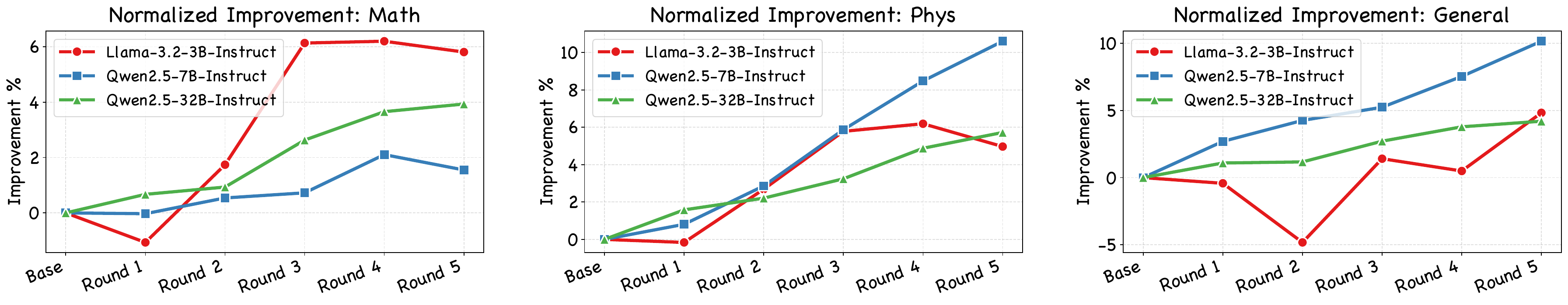}
    \caption{Normalized improvement trends relative to the base model across five training rounds. The results highlight the robust scalability of AERO across diverse model families (Llama-3.2, Qwen2.5) and parameter scales (3B to 32B) within the Mathematics, Physics, and General Reasoning domains.}
    \label{fig:trends}
\end{figure*}

We compare AERO with existing competitive self-evolving methods~\cite{huang2025r,zhao2025absolute}. Based on the results presented in Table~\ref{tab:main_v2}, we draw the following key insights.

\textbf{Superiority Over Competitive Baselines.} AERO demonstrates substantial superiority over competitive baselines across multiple reasoning domains. Most notably, on the AMC mathematical dataset using the Qwen3-4B-Base architecture, AERO R5 achieves a performance of 54.5\%, representing a significant \textbf{4.5\%} lead over the strongest baseline method, Absolute Zero (50.0\%). This performance advantage is sustained across most evaluation benchmarks, where AERO consistently provides higher reasoning accuracy than other data-free self-evolving methods.

\textbf{Effectiveness of Autonomous Reasoning Evolution.} The framework exhibits exceptional effectiveness in driving autonomous reasoning evolution compared to the base models. Specifically, Qwen3-4B-Base and Qwen3-8B-Base achieve average performance improvements of 4.6\% and 5.1\%, respectively, across the nine benchmarks. The evolution is particularly pronounced in the mathematical reasoning domain, where the average performance increases by \textbf{6.1\%} for the 4B model and \textbf{7.2\%} for the 8B model. Across most benchmarks, the LLM's reasoning capabilities evolve continuously through each training round, generally reaching their optimal performance in the final iteration. This steady growth confirms that AERO can successfully internalize advanced reasoning capabilities through purely endogenous feedback loops without relying on any external supervision.

\subsection{Analysis} \label{sec:analysis}
In this section, we conduct further experimental analysis to provide a comprehensive evaluation of AERO.

\textbf{Framework robustness.} To assess our framework's robustness, we apply AERO to various model families and sizes, including Llama-3.2-3B-Instruct, Qwen2.5-7B-Instruct, and Qwen2.5-32B-Instruct. As illustrated in Figure~\ref{fig:trends}, which visualizes the relative improvement of each round compared to the base model, all evaluated models demonstrate a clear and sustained upward trend across the three reasoning domains. Notably, even the smaller Llama-3.2-3B-Instruct model shows significant progress, maintaining a positive growth trajectory. Although showing persistent growth, the evolution eventually reaches saturation in the later stages, particularly for base models with smaller parameter scales; a detailed discussion of this saturation phenomenon is provided in Appendix~\ref{app:evolutionary-saturation}. These results confirm that AERO's endogenous feedback loops are not architecture-specific and can robustly drive the evolution of reasoning capabilities.

\begin{figure*}[ht]
    \centering
    \includegraphics[width=\textwidth]{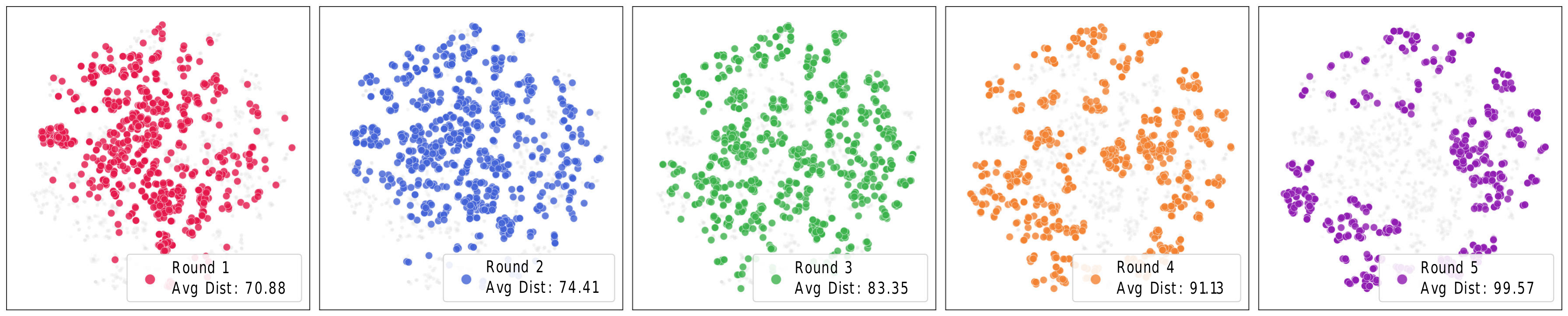}
    \caption{Manifold visualization via t-SNE~\cite{maaten2008visualizing} illustrates the evolutionary process of tasks synthesized across five rounds for Qwen2.5-7B-Instruct. The ``Avg Dist'' metric in each legend indicates the average pairwise Euclidean distance in the 2D latent space, serving as a proxy for task diversity within each round.}
    \label{fig:tsne}
\end{figure*}
\textbf{Ablation Study.} We conduct a comprehensive ablation study to quantify the individual contributions of internalizing three specialized functional capacities and the two core mechanisms within the AERO framework.
\begin{table}[t]
\small
\caption{Ablation Study. All reported performance metrics represent the best results achieved across the five rounds. Superscripts indicate performance drops compared with standard AERO.}
\label{tab:ablation}
\setlength{\tabcolsep}{2pt}

\newcommand{\drop}[1]{\textsuperscript{-#1}}

\begin{tabular}{@{}lcccc@{}}
\toprule
\multicolumn{1}{c}{Method} & Overall & Math AVG & Phys AVG & General AVG \\ 
\midrule
\multicolumn{5}{l}{Qwen2.5-7B-Instruct} \\ 
\midrule
Base Model                 
& 45.7          
& 70.3          
& 27.3          
& 39.7          \\

\rowcolor[HTML]{EFEFEF} 
AERO                       
& \textbf{48.5} 
& \textbf{71.8} 
& \textbf{30.0} 
& \textbf{43.7} \\ 
\midrule

w/o Self-Question          
& 46.0\drop{2.5} 
& 70.3\drop{1.5} 
& 27.8\drop{2.2} 
& 42.1\drop{1.6} \\

w/o Self-Answer            
& 45.8\drop{2.7} 
& 70.3\drop{1.5} 
& 28.2\drop{1.8} 
& 41.1\drop{2.6} \\

w/o Self-Critisim          
& {\ul 48.1}\drop{0.4} 
& {\ul 71.5}\drop{0.3} 
& {\ul 29.5}\drop{0.5} 
& {\ul 43.1}\drop{0.6} \\ 
\midrule

w/o ZPD                    
& 45.8\drop{2.7} 
& 69.2\drop{2.6} 
& 28.4\drop{1.6} 
& 39.9\drop{3.8} \\

w/o ICC                    
& 46.9\drop{1.6} 
& 70.8\drop{1.0} 
& 28.1\drop{1.9} 
& 41.8\drop{1.9} \\ 
\bottomrule
\end{tabular}
\vskip -0.1in
\end{table}

Regarding the internalization of specialized roles' capacities, we implement the removal of specific capabilities by excluding the corresponding preference datasets $D_g^t$, $D_s^{t-1}$, or $D_r^{t-1}$ from the total optimization objective. As shown in Table~\ref{tab:ablation}, removing the internalization of either the Self-Questioning or Self-Answering capabilities leads to a substantial performance drop, with overall scores falling to 46.0 and 45.8, respectively, nearly regressing to the base model's performance (45.7). This indicates that the core evolutionary drive stems from internalizing the synergy between ZPD task synthesis and high-confidence reasoning. In contrast, excluding the internalization of Self-Criticism results in a more moderate decline to 48.06. These results suggest that while the internalization of the Generator-Solver loop provides the fundamental engine for capability growth, internalizing the Refiner role provides essential critical abilities necessary to achieve optimal results across all reasoning domains.

Beyond the internalization of specialized roles, we evaluate the necessity of our two core mechanisms. We first ablate the ZPD positioning by removing the entropy-based filtering process. Under this configuration, the LLM is trained on all synthesized tasks regardless of their difficulty level. As shown in Table~\ref{tab:ablation}, this leads to a significant performance decline to 45.8 (-2.7). This result proves that without ZPD positioning, the model falls into a sub-optimal learning zone that limits the growth of reasoning capabilities. Furthermore, we replace the ICC mechanism with a standard majority voting baseline to evaluate the importance of logic-based verification. While majority voting relies entirely on statistical consensus, ICC forces the LLM to perform convergence verification through independent reasoning paths. The drop in the overall score to 46.9 (-1.6) confirms that simple consensus is insufficient for providing the highly reliable feedback required for stable self-evolution. Collectively, these findings demonstrate that each component within AERO is essential to sustain effective self-evolution.

\begin{figure}[h]
    \centering
    \includegraphics[width=0.95\linewidth]{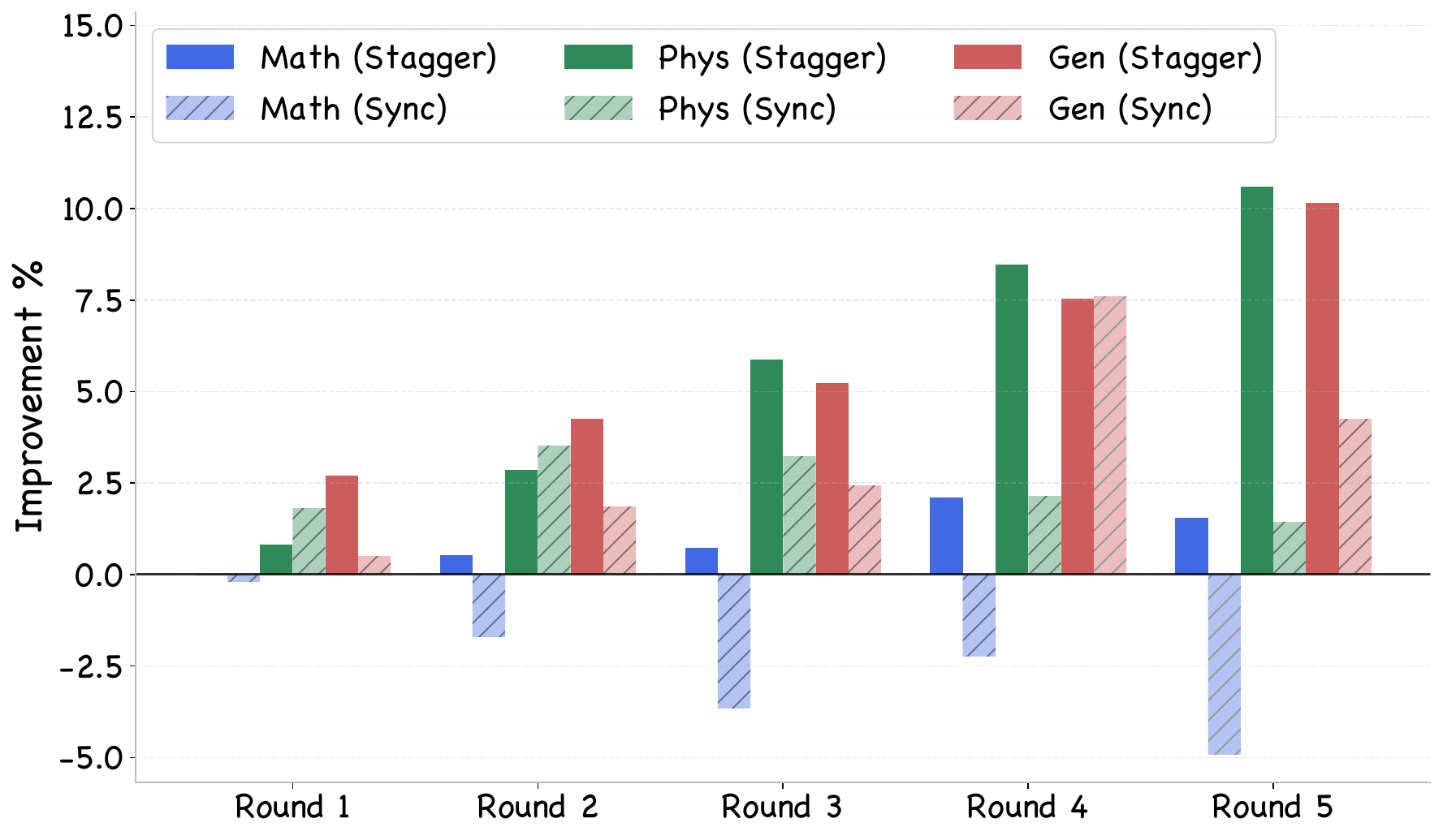}
    \caption{Comparison of staggered (solid) and synchronous (hatched) training performance. The results report the improvement rate of each round relative to the base model across three reasoning domains.}
    \label{fig:Stagger_vs_Sync_comp}
\end{figure}

\textbf{Effectiveness of Staggered Training Strategy.} 
We evaluate the effectiveness of our Staggered Training Strategy against a standard synchronous baseline using the Qwen2.5-7B-Instruct as the base model. As shown in Figure~\ref{fig:Stagger_vs_Sync_comp}, the synchronous strategy (Sync) suffers from curriculum collapse, where performance in domains like Mathematics even declines below the base model by the final rounds. By introducing a temporal offset, our staggered strategy synchronizes the development of questioning and solving roles. This ensures a steady upward trend in performance across Mathematics, Physics, and General Reasoning through 5 rounds, proving the strategy is essential for stable, long-term reasoning evolution.

\textbf{Reliability of ICC Pseudo-labels.}
We evaluate ICC precision by comparing its pseudo-labels against ground-truth data across five evolutionary rounds. Results indicate ICC maintains higher accuracy than traditional majority voting as tasks become increasingly complex. This advantage is particularly evident in later rounds, where statistical consensus reinforces collective hallucinations within the LLM. A detailed quantitative analysis of these results is provided in Appendix~\ref{app:pseudo-labels-reliability}.

\textbf{Qualitative Analysis.} To qualitatively evaluate the synthesized tasks' evolution process, we conduct a manifold visualization of the synthesized tasks' embeddings across five rounds for Qwen2.5-7B-Instruct in Figure~\ref{fig:tsne}. The Avg Dist metric shown in the legend tracks task diversity by calculating the average Euclidean distance between every pair of task points in the 2D space. As training progresses, the locations of these tasks show a clear and meaningful change. In the first two rounds, the task points are mostly clustered near the center. However, from Round 3 to 5, they spread out toward the outer edges and begin to form separate, specialized groups. This spatial expansion and the steady rise in Avg Dist values show that the LLM is actively exploring broader regions of the task space rather than merely exploiting repetitive patterns. Instead, it successfully finds more diverse areas to explore. Such a transition demonstrates that AERO effectively facilitates an automated curriculum where the generated tasks align with the LLM's improving reasoning capabilities. For a detailed description of the visualization pipeline, please refer to Appendix~\ref{app:manifold_visualization}.

\section{Conclusion}
We propose AERO, an unsupervised framework for autonomous reasoning evolution without expert-annotated data or external verifiers. By internalizing the synergistic roles of self-questioning, answering, and criticism, AERO enables a comprehensive improvement in reasoning capabilities. The integration of entropy-based ZPD positioning and ICC-based logical verification effectively targets the optimal learning zone and generates reliable feedback. Furthermore, the Staggered Training Strategy maintains evolutionary stability by synchronizing the growth across all functional roles. Extensive evaluations demonstrate that AERO effectively improves reasoning performance across multiple domains through purely endogenous feedback loops. Although AERO primarily targets tasks with well-defined answers, expanding AERO to open-ended domains like creative writing remains a valuable research direction. Overall, AERO establishes a scalable pathway for machine intelligence to autonomously evolve across complex reasoning frontiers.
\section*{Impact Statement}
This paper presents work whose goal is to advance the field of machine learning. There are many potential societal consequences of our work, none of which we feel must be specifically highlighted here.
\section*{Acknowledgments}
This work was supported by the National Key Research and Development Program of China (2022YFC3303600), the National Natural Science Foundation of China (62137002, 62306229,62477037, 62293553), the Key Research and Development Program of Shaanxi (2024GX-ZDCYL-02-12), the Youth Talent Support Program of Shaanxi Science and Technology Association (20240113), the China Post-doctoral Science Foundation (2024M752585, 2025T180425) and CAAI-Lenovo Blue Sky Research Fund (2025CAAI-LENOVO-06).
\bibliography{icml2026}

@article{guo2025deepseek,
  title={Deepseek-r1: Incentivizing reasoning capability in llms via reinforcement learning},
  author={Guo, Daya and Yang, Dejian and Zhang, Haowei and Song, Junxiao and Zhang, Ruoyu and Xu, Runxin and Zhu, Qihao and Ma, Shirong and Wang, Peiyi and Bi, Xiao and others},
  journal={arXiv preprint arXiv:2501.12948},
  year={2025}
}

@article{shao2024deepseekmath,
  title={Deepseekmath: Pushing the limits of mathematical reasoning in open language models},
  author={Shao, Zhihong and Wang, Peiyi and Zhu, Qihao and Xu, Runxin and Song, Junxiao and Bi, Xiao and Zhang, Haowei and Zhang, Mingchuan and Li, YK and Wu, Yang and others},
  journal={arXiv preprint arXiv:2402.03300},
  year={2024}
}

@article{ma2025deliberation,
  title={Deliberation on Priors: Trustworthy Reasoning of Large Language Models on Knowledge Graphs},
  author={Ma, Jie and Qu, Ning and Gao, Zhitao and Xing, Rui and Liu, Jun and Pei, Hongbin and Xie, Jiang and Song, Linyun and Wang, Pinghui and Tao, Jing and others},
  journal={arXiv preprint arXiv:2505.15210},
  year={2025}
}

@article{jaech2024openai,
  title={Openai o1 system card},
  author={Jaech, Aaron and Kalai, Adam and Lerer, Adam and Richardson, Adam and El-Kishky, Ahmed and Low, Aiden and Helyar, Alec and Madry, Aleksander and Beutel, Alex and Carney, Alex and others},
  journal={arXiv preprint arXiv:2412.16720},
  year={2024}
}

@article{lambert2024tulu,
  title={Tulu 3: Pushing frontiers in open language model post-training},
  author={Lambert, Nathan and Morrison, Jacob and Pyatkin, Valentina and Huang, Shengyi and Ivison, Hamish and Brahman, Faeze and Miranda, Lester James V and Liu, Alisa and Dziri, Nouha and Lyu, Shane and others},
  journal={arXiv preprint arXiv:2411.15124},
  year={2024}
}

@article{tao2024survey,
  title={A survey on self-evolution of large language models},
  author={Tao, Zhengwei and Lin, Ting-En and Chen, Xiancai and Li, Hangyu and Wu, Yuchuan and Li, Yongbin and Jin, Zhi and Huang, Fei and Tao, Dacheng and Zhou, Jingren},
  journal={arXiv preprint arXiv:2404.14387},
  year={2024}
}

@article{huang2025r,
  title={R-zero: Self-evolving reasoning llm from zero data},
  author={Huang, Chengsong and Yu, Wenhao and Wang, Xiaoyang and Zhang, Hongming and Li, Zongxia and Li, Ruosen and Huang, Jiaxin and Mi, Haitao and Yu, Dong},
  journal={arXiv preprint arXiv:2508.05004},
  year={2025}
}

@article{kuba2025language,
  title={Language self-play for data-free training},
  author={Kuba, Jakub Grudzien and Gu, Mengting and Ma, Qi and Tian, Yuandong and Mohan, Vijai and Chen, Jason},
  journal={arXiv preprint arXiv:2509.07414},
  year={2025}
}

@article{yu2025rlpr,
  title={RLPR: Extrapolating RLVR to General Domains without Verifiers},
  author={Yu, Tianyu and Ji, Bo and Wang, Shouli and Yao, Shu and Wang, Zefan and Cui, Ganqu and Yuan, Lifan and Ding, Ning and Yao, Yuan and Liu, Zhiyuan and others},
  journal={arXiv preprint arXiv:2506.18254},
  year={2025}
}

@inproceedings{tan2024large,
  title={Large Language Models for Data Annotation and Synthesis: A Survey},
  author={Tan, Zhen and Li, Dawei and Wang, Song and Beigi, Alimohammad and Jiang, Bohan and Bhattacharjee, Amrita and Karami, Mansooreh and Li, Jundong and Cheng, Lu and Liu, Huan},
  booktitle={EMNLP},
  year={2024}
}

@book{vygotsky1978mind,
  title={Mind in society: The development of higher psychological processes},
  author={Vygotsky, Lev S},
  volume={86},
  year={1978},
  publisher={Harvard university press}
}

@article{jin2025search,
  title={Search-r1: Training llms to reason and leverage search engines with reinforcement learning},
  author={Jin, Bowen and Zeng, Hansi and Yue, Zhenrui and Yoon, Jinsung and Arik, Sercan and Wang, Dong and Zamani, Hamed and Han, Jiawei},
  journal={arXiv preprint arXiv:2503.09516},
  year={2025}
}

@article{zheng2025deepeyes,
  title={DeepEyes: Incentivizing" Thinking with Images" via Reinforcement Learning},
  author={Zheng, Ziwei and Yang, Michael and Hong, Jack and Zhao, Chenxiao and Xu, Guohai and Yang, Le and Shen, Chao and Yu, Xing},
  journal={arXiv preprint arXiv:2505.14362},
  year={2025}
}

@article{lu2025ui,
  title={UI-R1: Enhancing Efficient Action Prediction of GUI Agents by Reinforcement Learning},
  author={Lu, Zhengxi and Chai, Yuxiang and Guo, Yaxuan and Yin, Xi and Liu, Liang and Wang, Hao and Xiao, Han and Ren, Shuai and Xiong, Guanjing and Li, Hongsheng},
  journal={arXiv preprint arXiv:2503.21620},
  year={2025}
}

@article{liu2025nover,
  title={NOVER: Incentive Training for Language Models via Verifier-Free Reinforcement Learning},
  author={Liu, Wei and Qi, Siya and Wang, Xinyu and Qian, Chen and Du, Yali and He, Yulan},
  journal={arXiv preprint arXiv:2505.16022},
  year={2025}
}

@article{zhou2025reinforcing,
  title={Reinforcing General Reasoning without Verifiers},
  author={Zhou, Xiangxin and Liu, Zichen and Sims, Anya and Wang, Haonan and Pang, Tianyu and Li, Chongxuan and Wang, Liang and Lin, Min and Du, Chao},
  journal={arXiv preprint arXiv:2505.21493},
  year={2025}
}

@inproceedings{yuan2024self,
  title={Self-rewarding language models},
  author={Yuan, Weizhe and Pang, Richard Yuanzhe and Cho, Kyunghyun and Li, Xian and Sukhbaatar, Sainbayar and Xu, Jing and Weston, Jason},
  booktitle={ICML},
  pages={57905--57923},
  year={2024}
}

@article{he2025visplay,
  title={VisPlay: Self-Evolving Vision-Language Models from Images},
  author={He, Yicheng and Huang, Chengsong and Li, Zongxia and Huang, Jiaxin and Yang, Yonghui},
  journal={arXiv preprint arXiv:2511.15661},
  year={2025}
}

@article{zhao2025absolute,
  title={Absolute zero: Reinforced self-play reasoning with zero data},
  author={Zhao, Andrew and Wu, Yiran and Yue, Yang and Wu, Tong and Xu, Quentin and Lin, Matthieu and Wang, Shenzhi and Wu, Qingyun and Zheng, Zilong and Huang, Gao},
  journal={arXiv preprint arXiv:2505.03335},
  year={2025}
}

@article{huang2025formarl,
  title={Formarl: Enhancing autoformalization with no labeled data},
  author={Huang, Yanxing and Jin, Xinling and Liang, Sijie and Li, Peng and Liu, Yang},
  journal={arXiv preprint arXiv:2508.18914},
  year={2025}
}

@article{chen2025r1,
  title={R1-Code-Interpreter: Training LLMs to Reason with Code via Supervised and Reinforcement Learning},
  author={Chen, Yongchao and Liu, Yueying and Zhou, Junwei and Hao, Yilun and Wang, Jingquan and Zhang, Yang and Fan, Chuchu},
  journal={arXiv preprint arXiv:2505.21668},
  year={2025}
}

@article{cai2025escaping,
  title={Escaping the Verifier: Learning to Reason via Demonstrations},
  author={Cai, Locke and Provilkov, Ivan},
  journal={arXiv preprint arXiv:2511.21667},
  year={2025}
}

@article{silver2017mastering,
  title={Mastering chess and shogi by self-play with a general reinforcement learning algorithm},
  author={Silver, David and Hubert, Thomas and Schrittwieser, Julian and Antonoglou, Ioannis and Lai, Matthew and Guez, Arthur and Lanctot, Marc and Sifre, Laurent and Kumaran, Dharshan and Graepel, Thore and others},
  journal={arXiv preprint arXiv:1712.01815},
  year={2017}
}

@article{sukhbaatar2017intrinsic,
  title={Intrinsic motivation and automatic curricula via asymmetric self-play},
  author={Sukhbaatar, Sainbayar and Lin, Zeming and Kostrikov, Ilya and Synnaeve, Gabriel and Szlam, Arthur and Fergus, Rob},
  journal={arXiv preprint arXiv:1703.05407},
  year={2017}
}

@inproceedings{chen2024self,
  title={Self-Play Fine-Tuning Converts Weak Language Models to Strong Language Models},
  author={Chen, Zixiang and Deng, Yihe and Yuan, Huizhuo and Ji, Kaixuan and Gu, Quanquan},
  booktitle={ICML},
  pages={6621--6642},
  year={2024},
  organization={PMLR}
}

@article{jiang2025bootstrapping,
  title={Bootstrapping task spaces for self-improvement},
  author={Jiang, Minqi and Lupu, Andrei and Bachrach, Yoram},
  journal={arXiv preprint arXiv:2509.04575},
  year={2025}
}

@article{wang2025space,
  title={SPACE: Noise contrastive estimation stabilizes self-play fine-tuning for large language models},
  author={Wang, Yibo and Chen, Qing-Guo and Xu, Zhao and Luo, Weihua and Zhang, Kaifu and Zhang, Lijun},
  journal={arXiv preprint arXiv:2512.07175},
  year={2025}
}

@article{ethayarajh2024kto,
  title={Kto: Model alignment as prospect theoretic optimization},
  author={Ethayarajh, Kawin and Xu, Winnie and Muennighoff, Niklas and Jurafsky, Dan and Kiela, Douwe},
  journal={arXiv preprint arXiv:2402.01306},
  year={2024}
}

@article{cobbe2021training,
  title={Training verifiers to solve math word problems},
  author={Cobbe, Karl and Kosaraju, Vineet and Bavarian, Mohammad and Chen, Mark and Jun, Heewoo and Kaiser, Lukasz and Plappert, Matthias and Tworek, Jerry and Hilton, Jacob and Nakano, Reiichiro and others},
  journal={arXiv preprint arXiv:2110.14168},
  year={2021}
}

@inproceedings{hendrycks2measuring,
  title={Measuring Mathematical Problem Solving With the MATH Dataset},
  author={Hendrycks, Dan and Burns, Collin and Kadavath, Saurav and Arora, Akul and Basart, Steven and Tang, Eric and Song, Dawn and Steinhardt, Jacob},
  booktitle={NeurIPS},
  year={2021}
}

@inproceedings{xuugphysics,
  title={UGPhysics: A Comprehensive Benchmark for Undergraduate Physics Reasoning with Large Language Models},
  author={Xu, Xin and Xu, Qiyun and Xiao, Tong and Chen, Tianhao and Yan, Yuchen and ZHANG, Jiaxin and Diao, Shizhe and Yang, Can and Wang, Yang},
  booktitle={ICML},
  year={2025}
}

@article{siddique2025physicseval,
  title={Physicseval: Inference-time techniques to improve the reasoning proficiency of large language models on physics problems},
  author={Siddique, Oshayer and Alam, JM and Rafy, Md Jobayer Rahman and Raiyan, Syed Rifat and Mahmud, Hasan and Hasan, Md Kamrul},
  journal={arXiv preprint arXiv:2508.00079},
  year={2025}
}

@article{qiu2025phybench,
  title={Phybench: Holistic evaluation of physical perception and reasoning in large language models},
  author={Qiu, Shi and Guo, Shaoyang and Song, Zhuo-Yang and Sun, Yunbo and Cai, Zeyu and Wei, Jiashen and Luo, Tianyu and Yin, Yixuan and Zhang, Haoxu and Hu, Yi and others},
  journal={arXiv preprint arXiv:2504.16074},
  year={2025}
}

@article{du2025supergpqa,
  title={Supergpqa: Scaling llm evaluation across 285 graduate disciplines},
  author={Du, Xinrun and Yao, Yifan and Ma, Kaijing and Wang, Bingli and Zheng, Tianyu and Zhu, King and Liu, Minghao and Liang, Yiming and Jin, Xiaolong and Wei, Zhenlin and others},
  journal={arXiv preprint arXiv:2502.14739},
  year={2025}
}

@article{wang2024mmlu,
  title={Mmlu-pro: A more robust and challenging multi-task language understanding benchmark},
  author={Wang, Yubo and Ma, Xueguang and Zhang, Ge and Ni, Yuansheng and Chandra, Abhranil and Guo, Shiguang and Ren, Weiming and Arulraj, Aaran and He, Xuan and Jiang, Ziyan and others},
  journal={NeurIPS},
  volume={37},
  pages={95266--95290},
  year={2024}
}

@inproceedings{rein2024gpqa,
  title={Gpqa: A graduate-level google-proof q\&a benchmark},
  author={Rein, David and Hou, Betty Li and Stickland, Asa Cooper and Petty, Jackson and Pang, Richard Yuanzhe and Dirani, Julien and Michael, Julian and Bowman, Samuel R},
  booktitle={COLM},
  year={2024}
}

@article{yang2025qwen3,
  title={Qwen3 technical report},
  author={Yang, An and Li, Anfeng and Yang, Baosong and Zhang, Beichen and Hui, Binyuan and Zheng, Bo and Yu, Bowen and Gao, Chang and Huang, Chengen and Lv, Chenxu and others},
  journal={arXiv preprint arXiv:2505.09388},
  year={2025}
}

@article{dubey2024llama,
  title={The llama 3 herd of models},
  author={Dubey, Abhimanyu and Jauhri, Abhinav and Pandey, Abhinav and Kadian, Abhishek and Al-Dahle, Ahmad and Letman, Aiesha and Mathur, Akhil and Schelten, Alan and Yang, Amy and Fan, Angela and others},
  journal={arXiv preprint arXiv:2407.21783},
  year={2024}
}

@article{qwen2025qwen25technicalreport,
      title={Qwen2.5 Technical Report}, 
      author={Qwen and : and An Yang and Baosong Yang and Beichen Zhang and Binyuan Hui and Bo Zheng and Bowen Yu and Chengyuan Li and Dayiheng Liu and Fei Huang and others},
      journal={arXiv preprint arXiv:2412.15115},
      year={2024}
}

@article{maaten2008visualizing,
  title={Visualizing data using t-SNE},
  author={Maaten, Laurens van der and Hinton, Geoffrey},
  journal={JMLR},
  volume={9},
  number={Nov},
  pages={2579--2605},
  year={2008}
}

@article{zhang2025interplay,
  title={On the interplay of pre-training, mid-training, and rl on reasoning language models},
  author={Zhang, Charlie and Neubig, Graham and Yue, Xiang},
  journal={arXiv preprint arXiv:2512.07783},
  year={2025}
}

@inproceedings{dongself,
  title={Self-play with Execution Feedback: Improving Instruction-following Capabilities of Large Language Models},
  author={Dong, Guanting and Lu, Keming and Li, Chengpeng and Xia, Tingyu and Yu, Bowen and Zhou, Chang and Zhou, Jingren},
  booktitle={ICLR},
  year={2025}
}

@article{li2025generalist,
  title={Generalist Reward Models: Found Inside Large Language Models},
  author={Li, Yi-Chen and Xu, Tian and Yu, Yang and Zhang, Xuqin and Chen, Xiong-Hui and Ling, Zhongxiang and Chao, Ningjing and Yuan, Lei and Zhou, Zhi-Hua},
  journal={arXiv preprint arXiv:2506.23235},
  year={2025}
}

@article{wang2025socratic,
  title={Socratic-zero: Bootstrapping reasoning via data-free agent co-evolution},
  author={Wang, Shaobo and Jiao, Zhengbo and Zhang, Zifan and Peng, Yilang and Ze, Xu and Yang, Boyu and Wang, Wei and Wei, Hu and Zhang, Linfeng},
  journal={arXiv preprint arXiv:2509.24726},
  year={2025}
}

@article{lin2025understanding,
  title={Understanding tool-integrated reasoning},
  author={Lin, Heng and Xu, Zhongwen},
  journal={arXiv preprint arXiv:2508.19201},
  year={2025}
}

@article{zelikman2022star,
  title={Star: Bootstrapping reasoning with reasoning},
  author={Zelikman, Eric and Wu, Yuhuai and Mu, Jesse and Goodman, Noah},
  journal={NeurIPS},
  volume={35},
  pages={15476--15488},
  year={2022}
}

@article{prabhudesai2025maximizing,
  title={Maximizing Confidence Alone Improves Reasoning},
  author={Prabhudesai, Mihir and Chen, Lili and Ippoliti, Alex and Fragkiadaki, Katerina and Liu, Hao and Pathak, Deepak},
  journal={arXiv preprint arXiv:2505.22660},
  year={2025}
}

@article{chen2025spc,
  title={Spc: Evolving self-play critic via adversarial games for llm reasoning},
  author={Chen, Jiaqi and Zhang, Bang and Ma, Ruotian and Wang, Peisong and Liang, Xiaodan and Tu, Zhaopeng and Li, Xiaolong and Wong, Kwan-Yee K},
  journal={arXiv preprint arXiv:2504.19162},
  year={2025}
}
\bibliographystyle{icml2026}

\newpage
\appendix
\onecolumn
\section{Related Work} \label{sec:related_word}
\textbf{Reinforcement Learning with Verifiable and Endogenous Rewards.} 
LLM reasoning has been significantly advanced by Reinforcement Learning from Verifiable Rewards (RLVR)~\cite{shao2024deepseekmath, lambert2024tulu, guo2025deepseek}, which provides deterministic feedback in specialized domains like mathematics and coding~\cite{zhao2025absolute, chen2025r1, huang2025formarl, jin2025search, zheng2025deepeyes, lu2025ui}. However, the applicability of RLVR is limited to fields where automated verifiers are available. To address this, recent research has shifted toward verifier-free paradigms using Endogenous feedback~\cite{li2025generalist}, such as decoding probabilities~\cite{yu2025rlpr} or self-generated likelihoods~\cite{zhou2025reinforcing, liu2025nover} as reward signals. While these methods attempt to incorporate expert demonstrations to guide the policy LLM~\cite{cai2025escaping}, they still rely on external data. In contrast, AERO achieves fully autonomous evolution by internalizing both generation and verification mechanisms, eliminating the need for external verifiers or expert supervision.

\textbf{Self-Evolving Language Models.} 
Inspired by the success of self-play in game-playing AI like AlphaZero~\cite{silver2017mastering, sukhbaatar2017intrinsic}, self-evolution allows LLMs to iteratively enhance their reasoning capabilities by learning from their own experiences~\cite{yuan2024self, tao2024survey, dongself, wang2025socratic}. While previous works focused on reward alignment~\cite{zelikman2022star,chen2024self, dongself}, the focus has shifted toward enhancing complex reasoning through ``challenger-solver" setups~\cite{zhao2025absolute, huang2025r, kuba2025language, lin2025understanding,chen2025spc}. Despite their promise, these methods face two major hurdles: they lack a principled way to calibrate task difficulty~\cite{kuba2025language} and often rely on majority voting or decoding confidence, which risks reinforcing incorrect priors and causing hallucinations~\cite{he2025visplay, huang2025r, prabhudesai2025maximizing}. AERO overcomes these limitations by introducing an entropy-guided ZPD mechanism for dynamic task selection and ICC for logic-based verification, creating a positive evolutionary cycle.
\section{Prompt} \label{sec: prompt}
\subsection{Generator}
\begin{promptBox}{Generator Prompt Template}
\begin{lstlisting}[
    breaklines=true,
    breakatwhitespace=true,
    breakindent=0pt,
    basicstyle=\ttfamily\small,
    columns=fullflexible
]
# Role
You are a Distinguished Professor in the Department of Physics and Mathematics, specializing in designing rigorous, competition-level theoretical problems. Your goal is to challenge advanced students with problems that prioritize deep conceptual insight and symbolic derivation over brute-force calculation.

# Task
Generate a medium-to-hard difficulty quantitative problem (Mathematics, Physics, or Theoretical Science) suitable for advanced undergraduates or early graduate students.

# Constraints
1. **Mathematical Rigor**: The problem must require clear definitions, precise assumptions, and logically sound derivations. Multi-step reasoning is mandatory.

2. **Quantitative Structure**: The problem should include well-defined parameters or constants. Calculations should lead to clean symbolic results, integers, or simple rational expressions. Avoid excessive numerical approximation unless conceptually necessary.

3. **Domain Variety**: Avoid repeatedly using the same domain.

4. **Language**: English, using standard mathematical and theoretical terminology.

5. **Formatting Constraints**: Output must be in **strict JSON format only**. No Markdown, explanations, or text outside the JSON object are allowed.

# JSON Structure
{
  "question": "A complete and self-contained mathematical or mathematical-physics problem statement, including all definitions, assumptions, and given constants.",
  "meta": {
    "knowledge_points": ["Key concept 1", "Key concept 2"],
    "domain": "Specify the primary domain",
    "background": "A brief 1-2 sentence description of the mathematical or physical context of the problem."
  }
}
\end{lstlisting}
\end{promptBox}
\subsection{Solver}
\begin{promptBox}{Solver Prompt Template}
\begin{lstlisting}[
    breaklines=true,
    breakatwhitespace=true,
    breakindent=0pt,
    basicstyle=\ttfamily\small,
    columns=fullflexible
]
# Role
You are a Senior Research Fellow with expertise in advanced quantitative sciences. Your task is to provide a "Gold Standard" solution that serves as a pedagogical reference for complex academic problems.

# Task
Execute a rigorous, step-by-step derivation for the provided problem, ensuring every logical transition is justified.

# Standardized Process
1. **Problem Analysis**: Identify the physical/mathematical framework and state all underlying assumptions.

2. **Symbolic Definition**: Explicitly define all variables, constants, and target unknowns using LaTeX.

3. **Analytical Derivation**: Construct the solution from first principles (laws, axioms, or theorems).

4. **Formal Computation**: Perform symbolic simplification or numerical evaluation with high precision.

5. **Final Synthesis**: State the final result clearly.

# Constraints
- Use LaTeX for ALL mathematical notation (e.g., $E = mc^2$).

- The final numerical or symbolic answer must be enclosed in \boxed{}.

# Problem
\end{lstlisting}
\end{promptBox}
\subsection{Refiner}
\begin{promptBox}{Refiner Prompt Template}
\begin{lstlisting}[
    breaklines=true,
    breakatwhitespace=true,
    breakindent=0pt,
    basicstyle=\ttfamily\small,
    columns=fullflexible
]
# Role
You are a rigorous Academic Reviewer with expertise in mathematics, physics, and quantitative sciences.  
You are provided with a **Problem** and a **Candidate Solution**, which is suspected to be **INCORRECT**.

# Task
1. Begin with the assumption that the Candidate Solution contains an error.

2. Carefully examine the logical reasoning, definitions, assumptions, derivations, and calculations.

3. Identify the precise flaw or unjustified step (the error may be subtle or conceptual).

4. **Re-solve the problem from first principles**, using a clear and logically sound approach.

5. Present the corrected result clearly, and **wrap the final answer in \\boxed{}** when an explicit result is required.

# Output Format
Thinking Process: <Analyze where the error or weakness occurs>
Correct Solution: <Complete and rigorous derivation or reasoning>
Final Answer: \\boxed{<Corrected result>}

# Input
\end{lstlisting}
\end{promptBox}

\subsection{Semantic Answer Clustering} \label{app:clustering_prompt}
\begin{promptBox}{Semantic Answer Equivalence Prompt Template}
\begin{lstlisting}[
    breaklines=true,
    breakatwhitespace=true,
    breakindent=0pt,
    basicstyle=\ttfamily\small,
    columns=fullflexible
]
# Role
You are a rigorous mathematical evaluator specializing in symbolic logic and quantitative equivalence.

# Task
Your goal is to determine whether the two provided expressions represent the same mathematical value or logical conclusion. You must account for different presentation formats, such as algebraic simplifications, numerical approximations, or symbolic variations.

# Instructions
1. Carefully analyze the underlying logic of Expression A and Expression B.
2. Determine if they are mathematically and logically identical regardless of their surface form.
3. Provide your final judgment as a structured JSON object.

Expression A: {expr_a}
Expression B: {expr_b}

# Output
Reply with strictly JSON: {{"equivalent": true}} or {{"equivalent": false}}.
\end{lstlisting}
\end{promptBox}
\section{Algorithm} \label{sec: algo}
Algorithm~\ref{alg:aero} formalizes the dual-loop optimization process, which is structured into two primary phases:
\begin{itemize}
    \item Experience Synthesis (Inner Loop, Lines 3–14): In the inner loop, the Generator synthesizes a batch of tasks $Q^t$. These tasks are filtered using entropy-based ZPD positioning, which calculates the normalized Shannon entropy of response clusters to identify the solvability gap where the model's current reasoning is neither trivial nor chaotic. For tasks within the optimal learning zone, the model performs Independent Counterfactual Correction, which leverages the Refiner role to verify reasoning paths, thereby producing high-fidelity endogenous labels without external ground truth. 
    \item Policy Optimization (Outer Loop, Lines 16–21): The outer loop leverages the Staggered Training Strategy to preserve evolutionary stability and prevent curriculum collapse. This approach ensures synchronized development across Self-Questioning, Self-Answering, and Self-Criticism capabilities, coordinating the growth of the LLM's diverse functional roles. Subsequently, the LLM parameters $\pi_\theta$ are updated via the KTO loss, which optimizes the policy based on binary feedback signals synthesized during the inner loop phase.
\end{itemize}
\begin{algorithm}[H]
\caption{Autonomous Evolutionary Reasoning Optimization (AERO)}
\label{alg:aero}
\begin{algorithmic}[1]
\REQUIRE Base model $\pi_\theta^{0}$; task batch size $m$; reasoning paths per task $n$; ZPD thresholds $[\tau_{low}, \tau_{high}]$; iterations $T$.
\FOR{$t = 1, \dots, T$}
    \STATE $\triangleright$ Inner Loop: Experience Synthesis
    \STATE $\mathcal{Q}^{t} = \{q_i\}_{i=1}^m \sim \pi_{\text{g}}^{t-1}$
    \FOR{each $q_i \in \mathcal{Q}^{t}$}
        \STATE $\mathcal{Y}_i^{t} = \{y_{i,j}^{t}\}_{j=1}^n \sim \pi_{\text{s}}^{t-1}(\cdot \mid q_i)$
        \STATE $\mathcal{C}_i^{t} \leftarrow \text{Cluster}(\mathcal{Y}_i^{t})$ \COMMENT{Group paths by the final extracted answers}
        \STATE $\bar{H}(q_i) \leftarrow \text{Entropy}(\mathcal{C}_i^{t})$ 
        \IF{$\bar{H}(q_i) \in [\tau_{low}, \tau_{high}]$}
            \STATE $\hat{y}_{i,1} \leftarrow \pi_{\text{r}}^{t-1}(\text{Correction} \mid q_i, \mathcal{C}_{i,1})$
            \STATE $\hat{y}_{i,2} \leftarrow \pi_{\text{r}}^{t-1}(\text{Correction} \mid q_i, \mathcal{C}_{i,2})$
            \STATE $\tilde{y}_{i}^{t} \leftarrow 
                \begin{cases} 
                \hat{y}_{i,1} & \text{if } \text{res}(\hat{y}_{i,1}) = \text{res}(\hat{y}_{i,2}) \\
                \perp & \text{otherwise}
                \end{cases}$
        \ENDIF
    \ENDFOR
    \STATE Construct $\mathcal{D}_{\text{g}}^{t}$, $\mathcal{D}_{\text{s}}^{t}$, and $\mathcal{D}_{\text{r}}^{t}$. (Eqs.~\ref{eq:d_g}--\ref{eq:d_r}).
    \STATE $\triangleright$ Outer Loop: Policy Optimization
    \IF{$t = 1$}
        \STATE $\mathcal{D}_{total}^{t} \leftarrow \mathcal{D}_{\text{g}}^{t}$
    \ELSE
        \STATE $\mathcal{D}_{total}^{t} \leftarrow \mathcal{D}_{\text{g}}^{t} \cup \mathcal{D}_{\text{s}}^{t-1} \cup \mathcal{D}_{\text{r}}^{t-1}$
    \ENDIF
    \STATE Update $\pi_\theta^{t-1}$ using $\mathcal{L}_{KTO}$ (Eq.~\ref{eq:kto_loss}) on $\mathcal{D}_{total}^{t} \rightarrow \pi_\theta^{t}$.
\ENDFOR

\STATE \textbf{return} $\pi_\theta^{T}$

\end{algorithmic}
\end{algorithm}
\section{Implementation Details} \label{implement_details}
\subsection{Semantic Equivalence Clustering} \label{clustering_details}
The computation of Normalized Shannon Entropy $\bar{H}(q_i^t)$ requires partitioning the $n$ reasoning trajectories in $\mathcal{Y}_i^t$ into $k$ unique clusters $\mathcal{C}_i^t$ based on their final answers. We denote $a_{i,j}^t$ as the symbolic or numerical answer extracted specifically from the content of the \texttt{\textbackslash boxed\{...\}} command within trajectory $y_{i,j}^t$. We define a binary equivalence function $f_{eq}(a_{i,j}^t, a_{i,l}^t) \in \{0, 1\}$ that utilizes the LLM-based judge described in Appendix~\ref{app:clustering_prompt} to evaluate whether two answers represent the same logical conclusion.

We implement a greedy one-pass clustering procedure to organize these trajectories. For each answer $a_{i,j}^t$ where $j$ ranges from 1 to $n$, the assignment to a cluster $c_{i,m}^t$ follows the update rule:
\begin{equation}
c_{i,m}^t \leftarrow c_{i,m}^t \cup \{y_{i,j}^t\} \quad \text{if} \quad \exists m \le k: f_{eq}(a_{i,j}^t, r_m) = 1,
\end{equation}
where $r_m$ is the representative seed of cluster $c_{i,m}^t$. If no such $m$ exists, a new cluster $c_{i,k+1}^t$ is initialized such that $r_{k+1} = a_{i,j}^t$. This process continues until all $n$ trajectories are processed. Trajectories that do not contain a \texttt{\textbackslash boxed\{...\}} tag or result in persistent parsing errors are assigned to a dedicated null cluster $c_{null}$ to maintain the integrity of the total sample size. 

Following the completion of the clustering process, the empirical probability for each semantic group is calculated as follows:
\begin{equation}
P(c_{i,j}^t) = \frac{|c_{i,j}^t|}{n}.
\end{equation}
This distribution forms the basis for calculating $\bar{H}(q_i^t)$, effectively mapping reasoning uncertainty to task difficulty. This clustering mechanism provides the necessary robust foundation for identifying the Zone of Proximal Development while maintaining computational efficiency during the iterative evolutionary process.
\subsection{Experimental Details} \label{exp_details}
The LLM is loaded from Hugging Face\footnote{\url{https://huggingface.co}} and trained using the LLaMA-Factory\footnote{\url{https://github.com/hiyouga/LLaMA-Factory}}. All training is conducted on eight NVIDIA H800-80GB GPUs, with bfloat16 precision enabled to reduce memory usage and accelerate training. We employed Low-Rank Adaptation (LoRA) across all linear layers with a rank of 16 to facilitate efficient adaptation during the KTO stage. The training process spanned 3 epochs, utilizing a cosine learning rate scheduler with an initial rate of $5.0 \times 10^{-6}$ and a 10\% warm-up ratio. With a per-device batch size of 1 and 8 gradient accumulation steps, the effective global batch size reached 64. For preference optimization, we applied a sigmoid loss function with a preference coefficient ($\beta$) of 0.1 and standard sample weights ($\lambda_p = \lambda_n = 1$).
\section{Evaluation Benchmarks} \label{benchmark}
\begin{itemize}
    \item GSM8K~\cite{cobbe2021training}: A dataset consisting of 8.5K high-quality grade school math word problems. It requires models to perform multi-step reasoning and serves as a standard benchmark for evaluating basic logical deduction and Chain-of-Thought (CoT) capabilities.
    \item MATH500~\cite{hendrycks2measuring}: A representative subset of 500 problems sampled from the MATH dataset (high school competition level). It covers diverse topics from algebra to calculus, specifically designed to test the model's ability to solve complex, competition-level mathematical problems.
    \item AMC: Comprising problems from the American Mathematics Competitions, this benchmark challenges models with tasks that require not only rigorous logical derivation but also mathematical intuition and creative problem-solving strategies.
    \item UGPhysics~\cite{xuugphysics}: Designed to evaluate undergraduate-level physics knowledge. It covers core curriculum areas such as classical mechanics, electromagnetism, thermodynamics, and quantum physics, assessing the model's mastery of advanced scientific concepts.
    \item PHYBench~\cite{qiu2025phybench}: A comprehensive physics benchmark focusing on complex scenario modeling. It evaluates a model’s proficiency in symbolic formula derivation, precise numerical calculation, and the deep interpretation of physical laws.
    \item PhysicsEval~\cite{siddique2025physicseval}: A multi-dimensional evaluation suite for physics literacy. It utilizes fine-grained task categorization to measure a model's performance in both conceptual differentiation and quantitative analysis.
    \item SuperGPQA~\cite{du2025supergpqa}: An expanded and enhanced version of the GPQA dataset. It includes a larger volume of high-difficulty questions authored by domain experts, spanning advanced fields in science, engineering, and medicine
    \item MMLU-Pro~\cite{wang2024mmlu}: An extension of the Massive Multitask Language Understanding (MMLU) benchmark. By increasing the number of choices and focusing on reasoning-intensive subjects, it provides a more discriminative and robust evaluation for state-of-the-art models.
    \item GPQA-Diamond~\cite{rein2024gpqa}: The most challenging subset of the Graduate-Level Google-Proof Q\&A (GPQA) dataset. These expert-written and verified questions are so difficult that even non-expert humans with access to the internet struggle to answer them correctly, making it a key metric for expert-level reasoning.
\end{itemize}

\section{Extended Analysis and Discussion}
\subsection{Reliability of ICC Pseudo-labels} \label{app:pseudo-labels-reliability}
To evaluate the reliability of the endogenous feedback signals, we compare the accuracy of pseudo-labels generated by Independent Counterfactual Correction (ICC) against the standard majority voting (MV) baseline using Qwen2.5-7B-Instruct as the base model. Since the synthesized tasks are generated autonomously and lack pre-existing ground-truth labels, we employ responses from DeepSeek-R1~\cite{guo2025deepseek} as proxy reference labels to facilitate this quantitative evaluation. Table~\ref{tab:pseudo_acc_appendix} presents the precision of pseudo-labels generated by ICC and MV. 

\begin{table}[h]
\centering
\small
\caption{Comparison of pseudo-label accuracy between Majority Voting and Independent Counterfactual Correction across five training rounds.}
\label{tab:pseudo_acc_appendix}
\begin{tabular}{@{}lccc@{}}
\toprule
\textbf{Round} & \textbf{MV Accuracy} & \textbf{ICC Accuracy} & \textbf{Improvement $\Delta$} \\ \midrule
Round 1        & 70.00\%             & 72.27\%              & +2.27\%                        \\
Round 2        & 62.32\%             & 74.97\%              & +12.65\%                       \\
Round 3        & 44.53\%             & 64.53\%              & +20.00\%                       \\
Round 4        & 43.28\%             & 61.04\%              & +17.76\%                       \\
Round 5        & 35.06\%             & 50.91\%              & +15.85\%                       \\ \bottomrule
\end{tabular}
\end{table}

The experimental results demonstrate that while the absolute accuracy of both methods declines as training progresses, Independent Counterfactual Correction consistently maintains a substantial lead over the majority voting baseline. This downward trend in accuracy is a natural consequence of the Generator moving toward more difficult reasoning frontiers within the Zone of Proximal Development. As pseudo-label accuracy declines sharply, the majority voting method becomes increasingly susceptible to collective hallucinations, reaching an accuracy of only 35.06 percent by the final round.

In contrast, our ICC method ensures that final feedback is based on internal logical consistency rather than simple statistical agreement. Starting from the second round, the performance gap between these two approaches widens significantly, reaching a peak improvement of 20.00\% in the third round. This trend confirms that ICC is substantially more robust than statistical consensus when handling complex reasoning tasks. Furthermore, the logical discrepancies identified throughout this process provide critical learning signals for the self-criticism functionality of the model. This ensures that the dual-loop evolutionary process remains driven by high-reliability feedback even in the absence of external labels.

\subsection{Evolution of Task Difficulty Distribution} \label{app:task-distribution-evolution}
\begin{figure}[h]
    \centering
    \includegraphics[width=\linewidth]{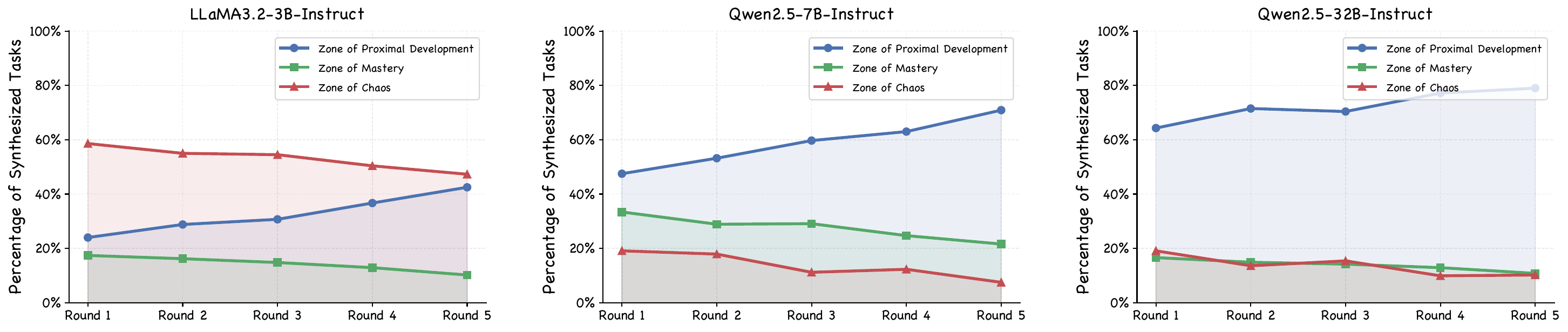}
    \caption{Evolution of synthesized task difficulty distribution over five evolutionary rounds. Tasks are categorized into the Zone of Mastery ($\bar{H}(q_i^t) < \tau_{low}$), Zone of Proximal Development ($\tau_{low} \le \bar{H}(q_i^t) \le \tau_{high}$), and Zone of Chaos ($\bar{H}(q_i^t) > \tau_{high}$) based on the normalized Shannon entropy of response clusters.}
    \label{fig:task_dist}
\end{figure}

To investigate how AERO facilitates automated curriculum learning, we analyze the distribution of synthesized tasks across three cognitive regions: the Zone of Mastery, the Zone of Proximal Development (ZPD), and the Zone of Chaos. Figure~\ref{fig:task_dist} illustrates the evolutionary trajectory of these distributions over five training rounds for LLaMA3.2-3B-Instruct, Qwen2.5-7B-Instruct, and Qwen2.5-32B-Instruct.

The primary observation is the robust and steady expansion of the ZPD across every base model architecture. For the Qwen2.5-32B-Instruct, the ZPD proportion increases from 64.3\% in Round 1 to 79.0\% by Round 5. Similarly, the Qwen2.5-7B-Instruct rises from 47.5\% to 70.9\%. Even for the smaller LLaMA3.2-3B-Instruct, the ZPD exhibits a significant expansion from 24.0\% to 42.5\% by the final round. This uniform upward trajectory demonstrates that AERO effectively enables the Generator to identify the solvability gap and target the difficulty interval with the highest learning efficiency, regardless of the base model's initial reasoning capacity or parameter scale.

The growth in ZPD is accompanied by a synchronized contraction of both the Zone of Mastery and the Zone of Chaos across all models. The consistent decline in the Zone of Mastery proves that AERO successfully avoids learning stagnation by filtering out tasks with negligible learning gradients. Simultaneously, the reduction in the Zone of Chaos indicates that as the model's capabilities advance, tasks once perceived as random noise are progressively internalized into the structured ZPD. While larger models exhibit higher initial precision in defining their reasoning boundaries, the fundamental mechanism of difficulty migration remains a universal property of the AERO dual-loop system, ensuring high learning pressure throughout the evolutionary process.
\subsection{Discussion on Evolutionary Saturation} \label{app:evolutionary-saturation}
The experimental results presented in Figure~\ref{fig:trends} indicate that the performance gains of AERO tend to reach a plateau or exhibit minor fluctuations by the fifth round, a phenomenon that is particularly evident in the 3B-parameter model. We attribute this evolutionary saturation to three primary factors.

One primary factor involves the inherent parameter constraints of LLaMA3.2-3B-Instruct, which impose a fundamental ceiling on its representative capacity for complex reasoning. As an unsupervised framework, AERO focuses on internalizing latent reasoning patterns already present within the pre-trained weights. Once these internal representations are fully refined, the model eventually reaches a saturation point in its ability to capture and represent increasingly sophisticated logical structures.

Furthermore, the solvability gap for smaller architectures appears both narrower and more fragile than that of larger models. While the proportion of tasks within the Zone of Proximal Development continues to grow throughout the evolutionary process, the 3B model struggles to maintain a dominant ZPD presence. As illustrated in Figure~\ref{fig:task_dist}, a significant portion of synthesized tasks either falls into the Zone of Chaos or remains in the Zone of Mastery, which restricts the availability of high-quality learning signals compared to larger-scale counterparts.

Additionally, the precision of endogenous feedback derived via Independent Counterfactual Correction, tends to diminish as the Generator shifts the task distribution toward more challenging frontiers. The quantitative analysis in Table~\ref{tab:pseudo_acc_appendix} confirms that pseudo-label accuracy declines as training progresses, although this trend remains more robust and reliable than baseline methods such as majority voting. For smaller models, this reduction in feedback fidelity leads to the accumulation of residual noise within the preference datasets, which ultimately destabilizes the policy optimization and hinders the model from transcending its current reasoning boundaries.

\subsection{Implementation of manifold visualization for Synthesized Tasks}\label{app:manifold_visualization}
To qualitatively assess the expansion of the reasoning frontier, we conduct a manifold visualization of the synthesized tasks across five evolutionary rounds. This analysis is performed using Qwen-2.5-7B-Instruct as the base model, where we visualize the entire set of $m = 1,000$ tasks synthesized in each round to ensure a comprehensive representation of the generative distribution. By projecting the high-dimensional semantic features of these 1,000 tasks into a two-dimensional plane, we can directly observe how the Generator explores the task space while maintaining the output within the Zone of Proximal Development.

The implementation of the visualization pipeline follows a three-stage process involving semantic encoding, non-linear projection, and diversity quantification. First, we transform the raw text of all 1,000 tasks into high-dimensional semantic embeddings. This is achieved using the paraphrase-multilingual-MiniLM-L12-v2 model\footnote{\url{https://huggingface.co/sentence-transformers/paraphrase-multilingual-MiniLM-L12-v2}}, which is a specialized Sentence-BERT architecture designed to map logical and mathematical text into a dense vector space where semantic similarity is preserved. 

Second, these high-dimensional embeddings are projected into a two-dimensional plane using t-Distributed Stochastic Neighbor Embedding, also referred to as t-SNE \cite{maaten2008visualizing}. To capture the local structural clusters within the task manifold, we set the perplexity to 10 and utilize PCA-based initialization to ensure stable projections across different rounds. Third, to provide a quantitative measure of the diversity of the synthesized tasks, we calculate the Average Euclidean Distance, denoted as $D_{avg}$, within the manifold space:
\begin{equation}
    D_{avg} = \frac{2}{m(m-1)} \sum_{1 \le i < j \le m} \| \mathbf{v}_i - \mathbf{v}_j \|_2.
\end{equation}
In this formalization, $\mathbf{v}_i$ and $\mathbf{v}_j$ represent the two-dimensional coordinate vectors of the tasks in the projected space, while $m$ remains fixed at 1,000 for all evaluations.

\section{Case Study} \label{case_study}
In this section, we present representative tasks sampled from the Zone of Proximal Development across five training rounds using Qwen2.5-7B-Instruct as the base model. These examples illustrate the evolutionary trajectory of AERO, showing how the generated tasks gradually transition to more complex and challenging scenarios that align with the model's advancing reasoning capabilities.

\begin{promptBox}{Round 1}
A uniform rod of length $L$ and mass $M$ is pivoted at one end, allowing it to oscillate freely within a vertical plane. We denote the angular displacement from the vertical axis as $\theta(t)$. Under the small-angle approximation where $\sin \theta \approx \theta$, our objective is to derive the governing equation of motion and determine the period of small oscillations. Furthermore, we shall compare this result to the period of a simple pendulum with a length equal to $L$ and identify the physical factors leading to any differences.
\end{promptBox}

\begin{promptBox}{Round 2}
Consider a function $f: \mathbb{R} \to \mathbb{R}$ defined by the functional equation:
\[
    f(x + y) = f(x)f(y) \quad \text{for all } x, y \in \mathbb{R}.
\]
Assume $f$ is not identically zero. Determine all possible forms of $f$ under these conditions and prove your assertion rigorously by considering the behavior of $f$ at specific points and using properties of exponential functions.
\end{promptBox}

\begin{promptBox}{Round 3}
Consider a sequence of functions $\{f_n(x)\}$ defined on the interval $[0,1]$, where
\[
f_n(x) = x^n(1-x)^n.
\]
Determine the limit
\[
\lim_{n \to \infty} \int_0^1 f_n(x)\,dx.
\]
Provide a rigorous proof of your result using appropriate convergence theorems and techniques from real analysis.
\end{promptBox}

\begin{promptBox}{Round 4}
A perfectly superconducting ring of mass $m$, radius $r$, and self-inductance $L$ is positioned in a horizontal plane above a fixed magnetic dipole. The dipole is located at the origin with its magnetic moment $\mathbf{m_0}$ aligned along the vertical $z$-axis. We assume the ring's radius $r$ is significantly smaller than its levitation height $h$ above the origin. At an initial height $z_0$, the ring carries zero current and is released from rest. As the ring moves under the influence of gravity and the magnetic field, it maintains its horizontal orientation.

\begin{enumerate}[label=(\alph*)]
    \item Determine the induced current $I$ in the superconducting ring as a function of its vertical position $z$ by applying the principle of magnetic flux conservation.
    \item Identify the equilibrium height $h$ where the magnetic levitation force exactly balances the gravitational force.
    \item Derive the governing equation for the vertical motion of the ring and calculate the frequency $\omega$ of small oscillations around the equilibrium position.
    \item Analyze the energy conversion process and find the maximum current $I_{max}$ flowing through the ring during its downward descent if it is released from a very large height.
\end{enumerate}
\end{promptBox}

\begin{promptBox}{Round 5}
A quantum harmonic oscillator is described by the Hamiltonian $H = \frac{p^2}{2m} + \frac{1}{2}m\omega^2x^2$, where $m$ is the mass, $\omega$ is the angular frequency, and $x$ and $p$ are the position and momentum operators. We let $|n\rangle$ denote the eigenstates of this system with corresponding energy eigenvalues $E_n = \hbar\omega(n + \frac{1}{2})$ for non-negative integers $n$. We define the lowering operator $a$ and the raising operator $a^\dagger$ such that they satisfy the commutation relation $[a, a^\dagger] = 1$ along with the eigenvalue equations $a|n\rangle = \sqrt{n}|n-1\rangle$ and $a^\dagger|n\rangle = \sqrt{n+1}|n+1\rangle$. 

\begin{enumerate}[label=(\alph*)]
    \item Prove that the action of the raising operator $a^\dagger$ on any eigenstate $|n\rangle$ increases its energy by exactly $\hbar\omega$.
    \item Using the ladder operator formalism, find the expectation value of the Hamiltonian $\langle n | H | n \rangle$ in terms of $n$ and $\hbar\omega$.
    \item Show that the uncertainty product $\sigma_x \sigma_p$ for the ground state $|0\rangle$ satisfies the Heisenberg Uncertainty Principle which states that $\sigma_x \sigma_p \ge \frac{\hbar}{2}$.
\end{enumerate}
\end{promptBox}



\end{document}